\documentclass[10pt,journal,compsoc]{IEEEtran}
\ifCLASSOPTIONcompsoc
  \usepackage[nocompress]{cite}
\else
  \usepackage{cite}
\fi
\ifCLASSINFOpdf
\else
\fi
\usepackage{amssymb}
\usepackage{amsmath}
\usepackage{amsfonts}
\usepackage{bm}
\usepackage{multirow}
\usepackage{bigstrut}
\usepackage{array}
\usepackage{gensymb}
\usepackage{pifont}
\usepackage{subfigure}
\usepackage{graphicx}
\usepackage{psfrag}
\usepackage{pstricks}
\usepackage{microtype}
\usepackage{enumitem}
\usepackage{comment}
\usepackage[ruled,linesnumbered]{algorithm2e}
\usepackage{makecell}
\usepackage{siunitx}
\usepackage{sidecap}
\usepackage[bookmarks,backref=true,linkcolor=black]{hyperref}
\hypersetup{
	pdfauthor = {},
	pdftitle = {},
	pdfsubject = {},
	pdfkeywords = {},
	colorlinks=true,
	linkcolor= black,
	citecolor= black,
	pageanchor=true,
	urlcolor = black,
	plainpages = false,
	linktocpage
}

\newcommand{\eigen}{\textsc{Eigen}}

\newcommand{\tabincell}[2]{\begin{tabular}{@{}#1@{}}#2\end{tabular}}

\DeclareMathOperator*{\argmin}{argmin}
\newcommand{\correspondence}{\emph{correspondence}}
\newcommand{\alignment}{\emph{alignment}}

\newcommand{\sourcept}{\mathbf{p}}
\newcommand{\targetpt}{\mathbf{q}}
\newcommand{\paringpt}{\widehat{\mathbf{q}}}
\newcommand{\paringnormal}{\widehat{\mathbf{n}}}

\newcommand{\sourceset}{P}
\newcommand{\targetset}{Q}
\newcommand{\numsourcept}{M}
\newcommand{\numtargetpt}{N}
\newcommand{\rmse}{r}
\newcommand{\homog}[1]{\tilde{#1}}
\newcommand{\liealg}[1]{\check{#1}}
\newcommand{\liealgparam}[1]{\widetilde{#1}}
\newcommand{\pplmap}{G_{\textrm{ppl}}}

\newcommand{\citemain}[1]{{\cite{#1}}}
\newcommand{\citeappx}[1]{{\cite{#1}}}

\usepackage{eufrak}
\newcommand{\seg}[1]{SE(#1)}
\newcommand{\sea}[1]{se(#1)}

\graphicspath{{Figs-ArXiv/}}

\SetAlFnt{\small}
\SetAlCapFnt{\small}
\SetAlCapNameFnt{\small}

\begin{document}
%
\title{Fast and Robust Iterative Closest Point}
%
%
%
%

\author{
	Juyong Zhang,~\IEEEmembership{Member,~IEEE,}~
	Yuxin Yao,~
	Bailin Deng$^\dagger$,~\IEEEmembership{Member,~IEEE}
	\IEEEcompsocitemizethanks{\IEEEcompsocthanksitem J. Zhang and Y. Yao are with School of Mathematical Sciences,
		University of Science and Technology of China.
		\IEEEcompsocthanksitem B. Deng is with  School of Computer Science and Informatics, Cardiff University.}
	\thanks{$^\dagger$Corresponding author. Email: \texttt{DengB3@cardiff.ac.uk}.}
}

\markboth{~}%
{~}
%


\IEEEtitleabstractindextext{%
\begin{abstract}
The Iterative Closest Point (ICP) algorithm and its variants are a fundamental technique for rigid registration between two point sets, with wide applications in different areas from robotics to 3D reconstruction. The main drawbacks for ICP are its slow convergence as well as its sensitivity to outliers, missing data, and partial overlaps. Recent work such as Sparse ICP achieves robustness via sparsity optimization at the cost of computational speed. In this paper, we propose a new method for robust registration with fast convergence.
First, we show that the classical point-to-point ICP can be treated as a majorization-minimization (MM) algorithm, and propose an Anderson acceleration approach to speed up its convergence.
In addition, we introduce a robust error metric based on the Welsch's function, which is minimized efficiently using the MM algorithm with Anderson acceleration. On challenging datasets with noises and partial overlaps, we achieve similar or better accuracy than Sparse ICP while being at least an order of magnitude faster.
Finally, we extend the robust formulation to point-to-plane ICP, and solve the resulting problem using a similar Anderson-accelerated MM strategy. Our robust ICP methods improve the registration accuracy on benchmark datasets while being competitive in computational time. 
\end{abstract}

\begin{IEEEkeywords}
Rigid Registration, Robust Estimator, Fixed-point iterations, Majorlazer Minimization method, Anderson Acceleration.
\end{IEEEkeywords}}

{\maketitle}

\IEEEdisplaynontitleabstractindextext

%
\IEEEpeerreviewmaketitle

\IEEEraisesectionheading{\section{Introduction}}
\label{sec:intro}

\IEEEPARstart{R}{igid} registration, which finds an optimal rigid transformation to align a source point set with a target point set, is a fundamental problem in computer vision and many other areas. 
The iterative Closest Point (ICP) algorithm~\citemain{BeslM92} is a classical method for rigid registration.
It alternates between closest point query in the target set and minimization of distance between corresponding points, and is guaranteed to converge to a locally optimal alignment. 
However, classical ICP can suffer from slow convergence due to its linear convergence rate~\citemain{PottmannHYH06}. Other registration methods have been developed with faster convergence. For example, in~\citemain{Chen1992} the alignment is performed by minimizing a point-to-plane distance, whereas in~\citemain{Pottmann2004} a locally quadratic approximant of the squared distance function is minimized. Both approaches are shown to have faster convergence rate than classical ICP~\citemain{PottmannHYH06}.
Another issue with ICP is that the alignment accuracy can be affected by imperfections of the point sets such as noises, outliers and partial overlaps, which often occur in real-world acquisition processes. Various techniques have been developed to address this problem. One popular approach is to disregard erroneous correspondence between points, using heuristics based on their distance or the angle between their normals~\citemain{RusinkiewiczL01}. Recently, an $\ell_p$-norm minimization approach is proposed in~\citemain{BouazizTP13} to induce sparsity of the distance between corresponding point pairs, which aligns the points in true correspondence while allowing large distance due to outliers and incomplete data. 

In this paper, we propose a novel and simple approach to address these two issues. Our key observation is that classical ICP is a majorization-minimization (MM) algorithm~\citemain{Lange2016} for minimizing $\ell_2$ distance between the two point sets, which iteratively constructs and minimizes a surrogate function and ensures monotonic decrease of the target energy. 
By treating this process as a fixed-point iteration, we speed up its convergence using 
\emph{Anderson acceleration} (AA)~\citemain{Walker2011}, an established numerical technique that proves effective for a variety of optimization problems in computer graphics~\citemain{PengDZGQL18}.  
In each iteration, Anderson acceleration computes an accelerated iterate based on the history of $m$ previous iterates.
Compared with existing approaches such as~\citemain{Chen1992,Pottmann2004}, our method does not require higher-order information such as normal or curvature which may not be available from the point cloud data and need to be estimated carefully in the presence of noise~\citemain{Mitra2003}.
Moreover, different from previous attempt on Anderson acceleration of ICP~\citemain{podto-aaicp-2017pre} that uses Euler angles to represent rotation, we adopt a parameterization of rigid transformation that does not suffer from singularity of Euler angles. Using the same MM framework, we can replace the squared distance metric used in classical ICP with a robust metric that is insensitive to noises, outliers, and partial overlaps. In particular, we adopt a robust metric based on the Welsch's function~\citemain{Holland77}, which allows for a simple quadratic surrogate function and can be minimized efficiently. Compared to the sparse ICP algorithm~\citemain{BouazizTP13}, our approach does not require introducing auxiliary variables for the solver, which leads to lower memory footprint and significantly faster convergence. 
We conduct a variety of experiments on both synthetic and real data, where our method improves the speed and robustness for alignment. 
Our approach can also be extended to other ICP formulations. In particular, we apply it to the point-to-plane ICP from~\citemain{Chen1992}, and achieve better registration accuracy than the original method on benchmark datasets. This illustrates the effectiveness of our method in improving robustness of ICP-type registration algorithms.

\begin{figure*}[!t]
	\centering
	\includegraphics[width=\textwidth]{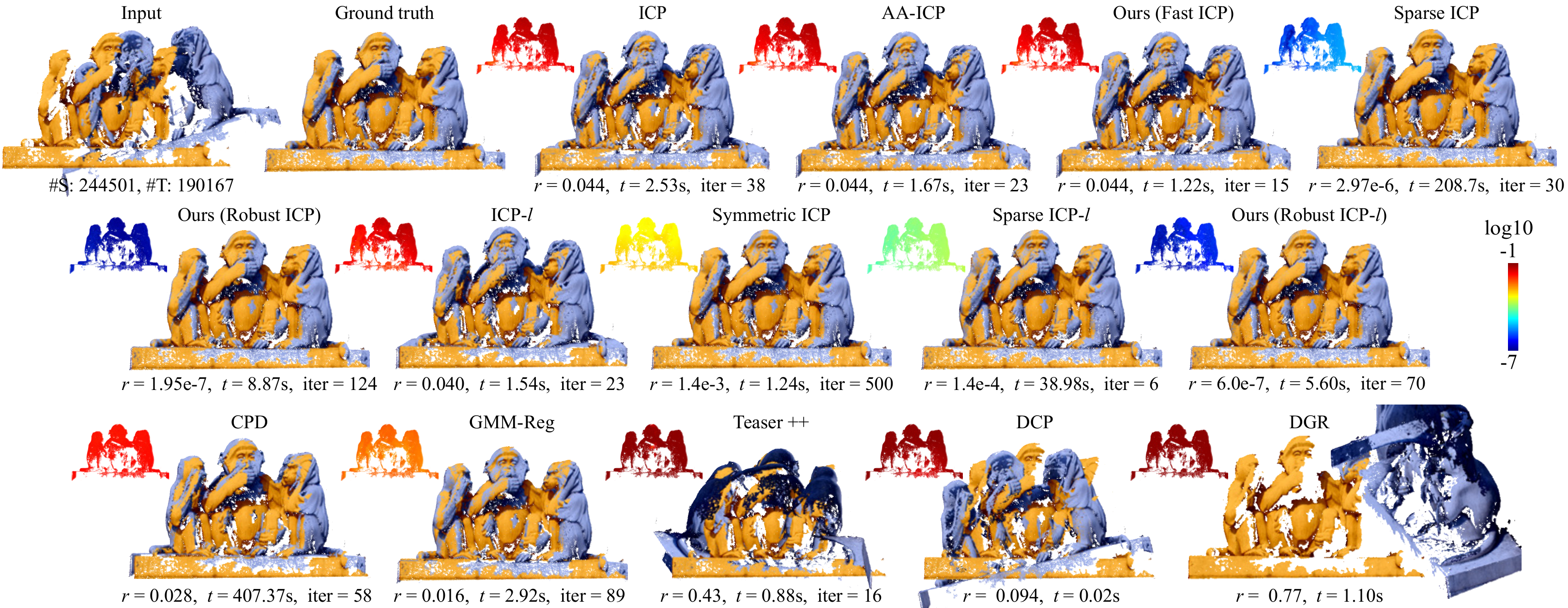}
	\caption{Comparison between different registration methods (see Section~\ref{sec:results}) on a pair of partially overlapping point clouds constructed using the monkey model from the EPFL statue dataset~\protect\citemain{EPFL2012}. \#S and \#T denote the number of points in the source and the target point clouds, respectively. Below each result we show the RMSE according to Eq.~\eqref{eq-mse}, the computational time, and the number of iterations. The log-scale color-coding illustrates the deviation between the transformed source point clouds using the computed alignment and the ground-truth alignment. Our robust point-to-point and point-to-plane ICP methods result in the lowest RMSE values, while being an order of magnitude faster than Sparse ICP.}
	\label{fig:teaser-monkeys}
\end{figure*}

To summarize, our main contributions include:
\begin{itemize}
	\item We propose a new formulation for Anderson-accelerated point-to-point ICP method. We parameterize rigid transformations via Lie algebra instead of Euler angles as in~\citemain{podto-aaicp-2017pre}, and use a more simple stabilization strategy than~\citemain{podto-aaicp-2017pre} that is easier to implement while guaranteeing monotonic decrease of the target energy. 
	\item We propose a robust metric for point-to-point alignment based on the Welsch's function, which is less sensitive to outliers and partial overlaps and can be solved efficiently using the MM framework with Anderson acceleration. Our method achieves similar or better registration accuracy than sparse ICP, while being significantly faster.
	\item We extend the formulation to point-to-plane ICP, using the Welsch's function to define a robust error metric that is minimized with an Anderson-accelerated MM solver. Our formulation improves the robustness of point-to-plane ICP without the need for point pair rejection.
\end{itemize} 
\section{Related Work}
\label{sec:related}

Registration is a classical research topic in computer vision and robotics due to its numerous applications such as 3D scene reconstruction and localization. For a comprehensive review of rigid and nonrigd registration, the reader is referred to~\citemain{TamCLLLMMSR13,TagliasacchiBPL16}. Here, we focus on ICP for rigid registration.
ICP and its variants~\citemain{BeslM92,Chen1992,RusinkiewiczL01,PomerleauCSM13,Rusinkiewicz19} start from an initial alignment, and alternate between correspondence update using closest-points lookup and alignment update based on the correspondence. Using this framework, an accurate registration relies on a good initial alignment as well as a robust way to update the alignment. 

For the initial alignment, Gelfand et al.~\citemain{Gelfand2005} computed shape descriptors on the point clouds, and used the descriptors to match feature points and determine a coarse alignment. Rusu et al.~\citemain{Rusu2009} performed similar matching using the Point Feature Histograms defined at each point. Aiger et al.~\citemain{Aiger2008} aligned two point clouds by matching a pair of co-planar 4-point sets from them that are approximately congruent. Later, Mellado et al.~\citemain{Mellado2014} proposed a more efficient approach for such alignment with linear time complexity.

To update the alignment, ICP minimizes the $\ell_2$ distance from the source points to their corresponding points~\citemain{BeslM92} or to the tangent planes at the corresponding points~\citemain{Chen1992}. Mitra et al.~\citemain{Mitra2004-Optimization} proposed a framework that determines the alignment by minimizing a squared distance function between the two point clouds, as well as a local quadratic approximant for efficient update of the alignment. A similar approach was taken in~\citemain{Pottmann2004} for aligning a point cloud to a surface. It was later shown in~\citemain{PottmannHYH06} that such a local quadratic approximant can lead to quadratic convergence. Recently, Rusinkiewicz~\citemain{Rusinkiewicz19} proposed a symmetrized objective function for ICP that yields faster convergence than point-to-point and point-to-plane ICP. Besides the convergence rate, another consideration for registration algorithms is their robustness to noises, outliers, and partial overlaps. A popular approach is to discard some point pairs from the alignment problem based on heuristics regarding their distance~\citemain{zhang1994iterative,RusinkiewiczL01,Chetverikov2005}. Other methods take a statistical approach and align two point sets via their Gaussian mixture representations~\citemain{Myronenko2010,Jian2011}. Another approach is to optimize a robust objective that reduces the influence from point pairs that are far apart~\citemain{Masuda1995,Trucco1999,Fitzgibbon03,BouazizTP13,ZhouPK16}. In~\citemain{BouazizTP13}, the objective is defined using the $\ell_p$-norm ($ p < 1$) to induce sparsity of the point-wise distances. Our robust metric is defined using Welsch's function instead, which also induces sparsity while allowing for a more efficient solver that guarantees convergence.

Besides ICP, other methods formulate registration as a global optimization problem~\citemain{Li2007-3D,Olsson2009,YangLCJ16}, which produces globally optimal results at the expense of higher computational costs.
In~\citemain{Yang2020}, a truncated least squares optimization is proposed to make the registration insensitive to outliers.
Recently, deep learning has also been applied to registration problems~\citemain{Wang2019,Choy2020}.

Anderson acceleration was originally proposed in~\citemain{Anderson1965} for iterative solution of nonlinear integral equations, and has proved effective for accelerating fixed-point iterations~\citemain{Eyert1996,Fang2009,Walker2011,Toth2015,Sterck2012,Lipnikov2013,Pratapa2016,Ho2017,Suryanarayana2019}. In computer graphics, Anderson acceleration has been applied recently to accelerate local-global solvers~\citemain{PengDZGQL18} and ADMM solvers~\citemain{Zhang2019-ADMM,Ouyang2020}. Classical Anderson acceleration can become unstable or stagnate~\citemain{Walker2011,Potra2013}. Peng et al.~\citemain{PengDZGQL18} proposed an stabilization strategy on optimization solvers based on the decrease of the target function. Recently, Anderson acceleration has been used in~\citemain{podto-aaicp-2017pre} to speed up the convergence of ICP. 
We also apply Anderson acceleration to ICP, but using a different representation of the transformation together with the stabilization strategy from~\citemain{PengDZGQL18}.

\section{Classical ICP Revisited}
\label{sec:pairing}

Given two sets of points $\sourceset = \{\sourcept_1, \ldots, \sourcept_\numsourcept\}$ and $\targetset = \{\targetpt_1, \ldots, \targetpt_\numtargetpt\}$ in $\mathbb{R}^d$, we optimize a rigid transformation on $\sourceset$ (represented using a rotation matrix $\mathbf{R} \in \mathbb{R}^{d \times d}$ and a translation vector $\mathbf{t} \in \mathbb{R}^d$) to align $\sourceset$ with $\targetset$:
\begin{equation}
	\min_{\mathbf{R}, \mathbf{t}} ~~ \sum_{i=1}^\numsourcept \bigl( D_i(\mathbf{R}, \mathbf{t}) \bigr)^2 + I_{SO(d)} (\mathbf{R}),
	\label{eq:ICPOptimization}
\end{equation}
where $D_i(\mathbf{R}, \mathbf{t}) = \min_{\targetpt \in \targetset} \|\mathbf{R} \sourcept_i + \mathbf{t} - \targetpt\|$ is the distance from the transformed point $\mathbf{R} \sourcept_i + \mathbf{t}$ to the target set $\targetset$,
and $I_{SO(d)}(\cdot)$ is an indicator function for the special orthogonal group $SO(d)$, which requires $\mathbf{R}$ to be a rotation matrix:
\begin{equation}
	I_{SO(d)}(\mathbf{R})
	=
	\left\{
	\begin{array}{ll}
	0, & \textrm{if}~ \mathbf{R}^T \mathbf{R} = \mathbf{I} ~\textrm{and}~\det(\mathbf{R}) = 1,\\
	+\infty, & \textrm{otherwise}.
	\end{array}
	\right.
\end{equation}
The ICP algorithm~\citemain{BeslM92} solves this problem using an iterative approach that alternates between the following two steps:
\begin{itemize}
	\item \emph{Correspondence step}: find the closest point $\paringpt_i^{(k)}$ in $\targetset$ for each point $\sourcept_i \in \sourceset$ based on  transformation $(\mathbf{R}^{(k)},\mathbf{t}^{(k)})$:
		\begin{equation}
			\paringpt_i^{(k)} = \argmin_{\targetpt \in \targetset} \left\| \mathbf{R}^{(k)} \sourcept_i + \mathbf{t}^{(k)} - \targetpt \right\|.
			\label{eq:CorresponcePoint}
		\end{equation}
	\item \emph{Alignment step}: update the transformation by minimizing the $\ell_2$ distance between the corresponding points:
		\begin{align}
			&(\mathbf{R}^{(k+1)}, \mathbf{t}^{(k+1)}) \nonumber\\
			= &\argmin_{\mathbf{R}, \mathbf{t}} \sum_{i=1}^\numsourcept \left\|\mathbf{R} \sourcept_i + \mathbf{t} - \paringpt_i^{(k)}\right\|^2 + I_{SO(d)} (\mathbf{R}).
			\label{eq:Alignment}
		\end{align}
\end{itemize}
The alignment step can be solved in closed form via SVD~\citemain{RigidSVD}.
This approach can be considered as a majorization-minimization (MM) algorithm~\citemain{Lange2004} for the problem~\eqref{eq:ICPOptimization}. To minimize a target function $f(x)$, each iteration of the MM algorithm constructs from the current iterate $x^{(k)}$ a surrogate function $g(x \mid x^{(k)})$ that bounds $f(x)$ from above, such that:
\begin{equation}
	f(x^{(k)}) = g(x^{(k)} \mid x^{(k)}),~
	\mathrm{and}~
	f(x) \leq g(x \mid x^{(k)})~\forall~x \neq x^{(k)}.
\label{eq:SurrogateFunc}
\end{equation}
The surrogate function is minimized to obtain the next iterate
\begin{equation}
	x^{(k+1)} = \argmin_{x} g(x \mid x^{(k)}).
	\label{eq:MMIteration}
\end{equation}
Equations~\eqref{eq:SurrogateFunc} and \eqref{eq:MMIteration} imply that
\[
	f(x^{(k+1)}) \leq g(x^{(k+1)} \mid x^{(k)}) \leq g(x^{(k)} \mid x^{(k)})
	= f(x^{(k)}).
\]
Therefore, the MM algorithm decreases the target function monotonically until it converges to a local minimum.
To see that ICP is indeed an MM algorithm, note that the target function for the alignment step is a surrogate function for the  target function in problem~\eqref{eq:ICPOptimization} and satisfies the conditions~\eqref{eq:SurrogateFunc}. Specifically, since the closest point $\paringpt_i^{(k)}$ is determined from $\mathbf{R}^{(k)}, \mathbf{t}^{(k)}$, we denote each distance value in~\eqref{eq:Alignment} as
\[
	d_i(\mathbf{R}, \mathbf{t} \mid \mathbf{R}^{(k)}, \mathbf{t}^{(k)}) = \left\|\mathbf{R} \sourcept_i + \mathbf{t} - \paringpt_i^{(k)}\right\|.
	\label{eq:IndividualSurrogate}
\]
Then from Eq.~\eqref{eq:CorresponcePoint} and the definition of $D_i$, we have
\[
	d_i(\mathbf{R}^{(k)}, \mathbf{t}^{(k)} \mid \mathbf{R}^{(k)}, \mathbf{t}^{(k)}) = D_i(\mathbf{R}^{(k)},\mathbf{t}^{(k)}).
\]
Moreover, from the definition of $D_i$, for any $\mathbf{R}, \mathbf{t}$:
\begin{align*}
	D_i(\mathbf{R} \sourcept_i + \mathbf{t}) &= \min_{\targetpt \in \targetset} \|\mathbf{R} \sourcept_i + \mathbf{t} - \targetpt\|\\
	&\leq \left\|\mathbf{R} \sourcept_i + \mathbf{t} - \paringpt_i^{(k)} \right\| = d_i(\mathbf{R}, \mathbf{t} \mid \mathbf{R}^{(k)}, \mathbf{t}^{(k)}).
\end{align*}
Thus each squared distance term in Eq.~\eqref{eq:Alignment} is a surrogate function for the corresponding term $\bigl( D_i(\mathbf{R}, \mathbf{t}) \bigr)^2$ in Eq.~\eqref{eq:ICPOptimization}, and the target function in Eq.~\eqref{eq:Alignment} is a surrogate function for the overall target function in Eq.~\eqref{eq:ICPOptimization} constructed from $\mathbf{R}^{(k)}$ and $\mathbf{t}^{(k)}$. Therefore,  ICP is an MM algorithm that decreases the target function of~\eqref{eq:ICPOptimization} monotonically until convergence.

\section{Fast and Robust ICP}
\label{sec:AA}

Despite its simplicity, classical ICP can be slow to converge to a local minimum due to its linear convergence rate~\citemain{PottmannHYH06}. 
In this section, we interpret ICP as a fixed-point iteration, and propose a method to improve its convergence rate using Anderson acceleration~\citemain{Anderson1965,Walker2011}, an established technique for accelerating fixed-point iterations.
In addition, classical ICP can lead to erroneous alignment in the presence of outliers and partial overlaps, due to the use of $\ell_2$ distance as the error metric in the alignment step. 
We adopt a robust error metric based on Welsch's function instead, and derive an MM solver for the resulting optimization problem, with Anderson acceleration to speed up its convergence.
In the following, we first review the basics of Anderson acceleration. 

\subsection{Preliminary: Anderson Acceleration}
\label{sec:AABasics}
Given a fixed-point iteration $x^{(k+1)} = G(x^{(k)})$, we define its \emph{residual} function as $F(x) = G(x) - x$, and denote $F^{(k)} = G(x^{(k)})$.
By definition, a fixed-point $x^\ast$ of the mapping $G(\cdot)$ satisfies $F(x^\ast) = 0$.
Anderson acceleration utilizes the latest iterate $x^{(k)}$ as well as its preceding $m$ iterates $x^{(k-m)}, x^{(k-m+1)}, \ldots, x^{(k-1)}$ to derive a new iterate $x_{\textrm{AA}}^{(k+1)}$ that convergences faster to a fixed point~\citemain{Walker2011}:
\begin{equation}
	x_{\textrm{AA}}^{(k+1)} = G(x^{(k)}) - \sum_{j=1}^m \theta_j^\ast \bigl(G(x^{(k-j+1)}) - G(x^{(k-j)})\bigr),
	\label{eq:AndersonAcceleration}
\end{equation}
where $(\theta_1^\ast, \ldots, \theta_m^\ast)$ is the solution to the following linear least-squares problem:
\begin{equation*}
(\theta_1^\ast, \ldots, \theta_m^\ast) = \argmin
\Bigl\| F^{(k)} - \sum_{j=1}^m \theta_j \bigl(F^{(k-j+1)} - F^{(k-j)}\bigr) \Bigr\|^2,
\end{equation*}
It has been shown that Anderson acceleration is a quasi-Newton method for finding a root of the residual function~\citemain{Fang2009}, 
and it can improve the convergence rate for fixed-point iterations that converge linearly~\citemain{evans2020proof}. 

\subsection{Applying Anderson Acceleration to ICP}
\label{eq:AAForICP}
\begin{figure}[!t]
	\centering
	\includegraphics[width=0.85\columnwidth]{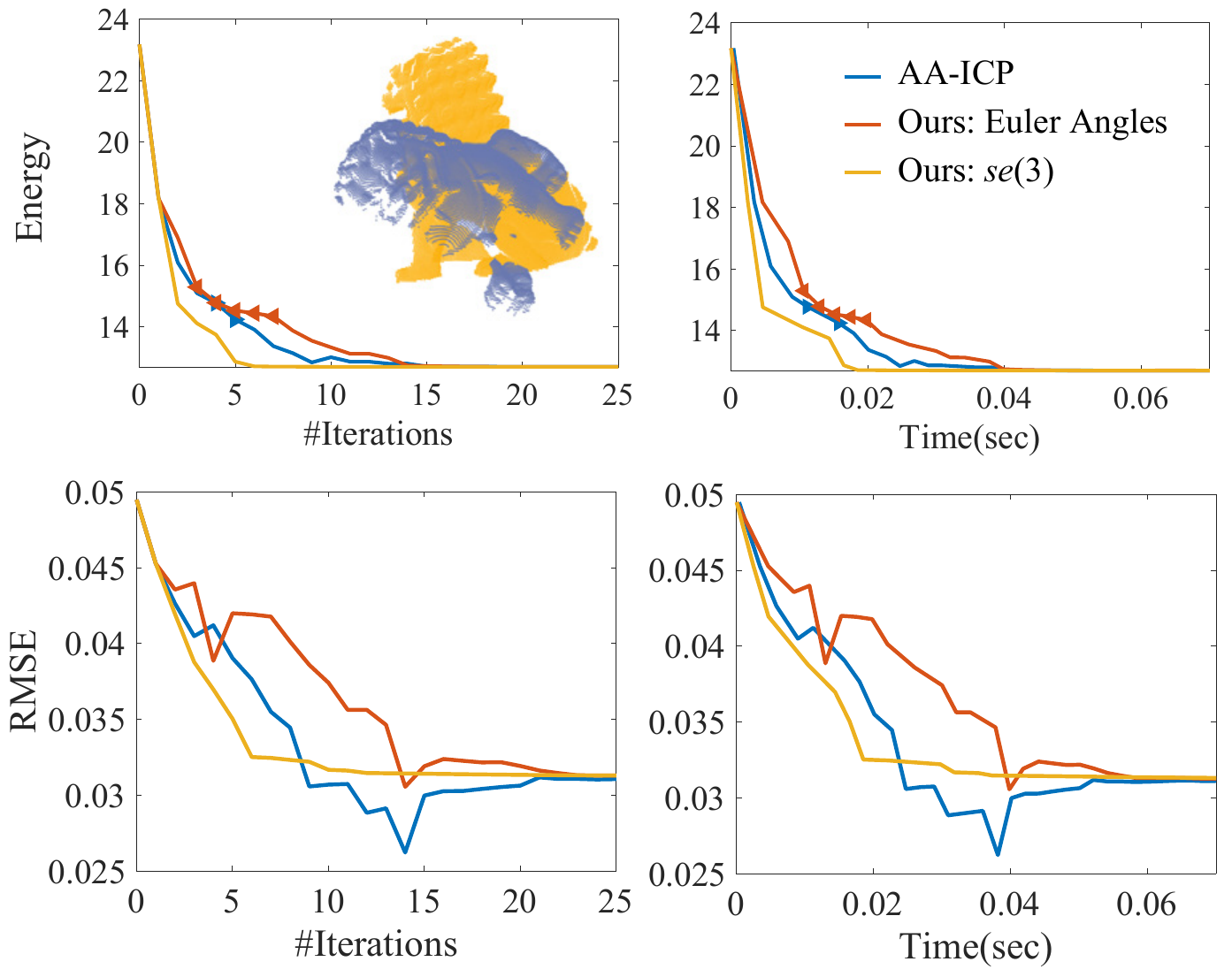}
	\caption{Target energy and RMSE plots for Anderson-accelerated ICP methods on a pair of point clouds, using different transformation representations and stabilization strategies. Our formulation outperforms AA-ICP~\protect\citemain{podto-aaicp-2017pre} as well as a Euler angle-based method using our stabilization strategy. For Euler angle-based methods, we use solid triangle symbols to highlight the iterates that are close to the gimbal lock.}
	\label{fig:Dragon_euler}
\end{figure}

The classical ICP explained in Section~\ref{sec:pairing} can be written as a fixed-point iteration of the transformation variables $\mathbf{R}$ and $\mathbf{t}$:
\begin{equation}
	(\mathbf{R}^{(k+1)}, \mathbf{t}^{(k+1)}) =  G_{\textrm{ICP}}(\mathbf{R}^{(k)}, \mathbf{t}^{(k)}),
	\label{eq:FixedPointICP}
\end{equation}
where
\begin{align*}
	& G_{\textrm{ICP}}(\mathbf{R}^{(k)}, \mathbf{t}^{(k)})\\
	=& \argmin_{\mathbf{R}, \mathbf{t}} \sum_{i=1}^\numsourcept \left\|\mathbf{R} \sourcept_i + \mathbf{t} - \Pi_{Q}(\mathbf{R}^{(k)} \sourcept_i + \mathbf{t}^{(k)}) \right\|^2 + I_{SO(d)} (\mathbf{R}),
\end{align*}
and $\Pi_{Q}(\cdot)$ denotes the closest projection onto the point set $Q$. 
However, we cannot directly apply Anderson acceleration to the mapping $G_{\textrm{ICP}}$. This is because Anderson acceleration will compute the new value of $\mathbf{R}$ as an affine combination of rotation matrices, which is in general not a rotation matrix itself. 
To address this issue, we can parameterize a rigid transformation using another set of variables $\mathbf{X}$, such that any value of $\mathbf{X}$ corresponds to a valid rigid transformation, and the ICP iteration can be re-written in the form of
\begin{equation}
 \mathbf{X}^{(k+1)} =  \overline{G}_{\textrm{ICP}}(\mathbf{X}^{(k)}).
\end{equation}
Then we can apply Anderson acceleration to the variable $\mathbf{X}$ by performing the following steps in each iteration:
\begin{enumerate}
	\item From the current variable $\mathbf{X}^{(k)}$, recover the rotation matrix $\mathbf{R}^{(k)}$ and translation vector $\mathbf{t}^{(k)}$.
	\item Perform the ICP update $(\mathbf{R'}, \mathbf{t'}) = G_{\textrm{ICP}}(\mathbf{R}^{(k)}, \mathbf{t}^{(k)})$.
	\item Compute the parameterization of $(\mathbf{R'}, \mathbf{t'})$ to obtain $\overline{G}_{\textrm{ICP}}(\mathbf{X}^{(k)})$.
	\item Compute the accelerated value $\mathbf{X}_{\textrm{AA}}$ with Eq.~\eqref{eq:AndersonAcceleration} using $\mathbf{X}^{(k-m)}, \ldots, \mathbf{X}^{(k)}$ and $G_{\textrm{ICP}}(\mathbf{X}^{(k-m)}), \ldots, G_{\textrm{ICP}}(\mathbf{X}^{(k)})$.
\end{enumerate}
One possible parameterization of rigid transformations is to concatenate the translation vector and the Euler angles of the rotation~\citemain{Diebel2006,Hoffman1972}. This is the approach taken by the AA-ICP method~\citemain{podto-aaicp-2017pre} for applying Anderson acceleration to  ICP in $\mathbb{R}^3$. 
However, it is well known that the Euler angle representation has singularities called the \emph{gimbal lock}~\citemain{Diebel2006}. This can affect the performance of AA-ICP when the optimal rotation is close to a gimbal lock (see Fig.~\ref{fig:Dragon_euler} for an example). 
An alternative representation of rotation in $\mathbb{R}^3$ without such singularities is the unit quaternions, which are identified with unit vectors in $\mathbf{R}^4$~\citemain{Diebel2006}.
This representation is not suitable either, as an affine combination of unit vectors does not result in a unit vector in general.
Rather than the using the above representations, we note that all rigid transformations in $\mathbb{R}^d$ form the \emph{special Euclidean group} $\seg{d}$, which is a \emph{Lie group} and gives rise to a \emph{Lie algebra} $\sea{d}$ that is a vector space~\citemain{Varadarajan1984}. From a differential geometry perspective, $\seg{d}$ is a smooth manifold and $\sea{d}$ is its tangent space at the identity transformation. We can then parameterize rigid transformations using their corresponding elements in $\sea{d}$. 

Specifically, if we represent each point $\mathbf{p} \in \mathbb{R}^d$ using its homogeneous coordinates
$
	\homog{\mathbf{p}} =
	[
	\mathbf{p}^T,
	1
	]^T
$,
then a rigid transformation in $\mathbb{R}^d$ with rotation $\mathbf{R} \in \mathbb{R}^{d \times d}$ and translation $\mathbf{t} \in \mathbb{R}^d$ can be represented as a transformation matrix
$$
	\mathbf{T} =
	\begin{bmatrix}
	\mathbf{R} & \mathbf{t}\\
	\mathbf{0} & 1
	\end{bmatrix}
	\in \mathbb{R}^{(d+1) \times (d+1)}
$$
for the homogeneous coordinates.
All such matrices form the special Euclidean group $\seg{d}$. Its Lie algebra $\sea{d}$ contains matrices of the following form
\begin{equation}
\liealg{\mathbf{T}} =
\begin{bmatrix}
\mathbf{S} & \mathbf{u}\\
\mathbf{0} & 0
\end{bmatrix}
\in \mathbb{R}^{(d+1) \times (d+1)},
\label{eq:Matrixsed}
\end{equation}
Each matrix $\liealg{\mathbf{T}} \in \sea{d}$ corresponds to a matrix $\mathbf{T} \in \seg{d}$ via the matrix exponential:
\begin{equation}
	\mathbf{T} = \exp\bigl(\liealg{\mathbf{T}}\bigr) = \sum_{i=0}^{\infty} \frac{1}{i!} \liealg{\mathbf{T}}^{i}.
	\label{eq:MatrixExp}
\end{equation}
The matrix exponential can be computed numerically using a generalization of Rodrigues' method~\citemain{Gallier2002Computing}.
On the other hand, given a matrix $\mathbf{T} \in \seg{d}$, there may be more than one matrix $\liealg{\mathbf{T}} \in \sea{d}$ that satisfies Eq.~\eqref{eq:MatrixExp}.
In Appendix~\ref{sec:MatrixLog}, we present a method to determine a unique value of $\liealg{\mathbf{T}}$. 
We call it the \emph{logarithm} of $\mathbf{T}$, and denote it by
$\liealg{\mathbf{T}} = \log(\mathbf{T})$.
We then perform Anderson acceleration on the logarithms of the transformations.
Since $\sea{d}$ is a vector space, the accelerated value $\liealg{\mathbf{T}}_{\textrm{AA}}$\textemdash which is computed as an affine combination of elements in $\sea{d}$\textemdash also belongs to $\sea{d}$ and represents a rigid transformation ${\mathbf{T}}_{\textrm{AA}} = \exp\bigl(\liealg{\mathbf{T}}_{\textrm{AA}}\bigr) \in \seg{d}$. 

Simply applying Anderson acceleration as explained in Section~\ref{sec:AABasics} is often not sufficient for fast convergence. It is known that Anderson acceleration can suffer from instability and stagnation even for linear problems~\citemain{Potra2013}, thus safeguarding steps are often necessary to improve its performance~\citemain{PengDZGQL18,fu2019anderson,Zhang2019-ADMM}. To this end, we follow the stabilization strategy proposed in~\citemain{PengDZGQL18}: we accept the accelerated value as the new iterate only if it decreases the target function~\eqref{eq:ICPOptimization} compared with the previous iterate; otherwise, we revert to the un-accelerated ICP iterate as the new iterate.
This approach is more simple than the combination of heuristics in~\citemain{podto-aaicp-2017pre}, while ensuring monotonic decrease of the target energy.
Following~\citemain{PengDZGQL18}, we set the number of previous iterates for Anderson acceleration to $m = 5$ in all experiments.

Compared to AA-ICP~\citemain{podto-aaicp-2017pre} that also applies Anderson acceleration to ICP, our approach differs in two aspects. First, we apply Anderson acceleration via the Lie algebra $\sea{d}$ instead of the Euler angles, which is free from the singularities of gimbal locks. Second, our stabilization strategy is more simple to implement than the multiple heuristics in~\citemain{podto-aaicp-2017pre} while ensuring monotonic decrease of the target function.
Fig.~\ref{fig:Dragon_euler} compares our method with AA-ICP, as well as an alternative Anderson acceleration approach using Euler angle representation and our stabilization strategy. The comparison is done on a synthetic model from~\citemain{ZhouPK16} for which the ground-truth alignment is known, and the point sets  are pre-aligned using Super4PCS~\citemain{Mellado2014}. 
We plot the value of target function~\eqref{eq:ICPOptimization} with respect to the iteration count and computational time, as well as the following root mean square error (RMSE) between the computed alignment $(\mathbf{R}, \mathbf{t})$ from the ground-truth alignment $(\mathbf{R}^{\ast}, \mathbf{t}^{\ast})$:
\begin{equation}
\rmse = \sqrt{\frac{1}{\numsourcept}\sum\nolimits_{i=1}^{\numsourcept}\|\mathbf{R}^{\ast}\sourcept_i + \mathbf{t}^{\ast} - \mathbf{R} \sourcept_i - \mathbf{t}\|_2^2}.
\label{eq-mse}
\end{equation}
Fig.~\ref{fig:Dragon_euler} shows that our method using the Lie algebra leads to faster convergence.
In the energy-iteration plots, for each Euler angle-based approach we use solid triangles to highlight the iterations that are close to the gimbal lock (with the pitch angle less than $0.01\pi$ away from $\pm \pi/2$).

\subsection{Robust ICP via Welsch's Function}
\label{sec:robust}
Classical ICP measures the alignment error using $\ell_2$ distance, which penalizes large deviation from \emph{any} point in the source set $\sourceset$ to the target set $\targetset$. This enables a closed-form solution in the alignment step, but may lead to erroneous alignment in the presence of outliers and partial overlaps: in such cases some points in $\sourceset$ may not correspond to any point in $\targetset$, and the ground-truth alignment can induce a large error that would be prohibited by the $\ell_2$ minimization. The issue can be resolved by adopting error metrics that promote {sparsity} of the point-wise distance from $\sourceset$ to $\targetset$. Such metrics penalize the distance between points in true correspondence, while allowing for large deviation induced by outliers and partial overlaps. One example is the $\ell_p$-norm of point-wise distance with $p \in (0, 1)$, resulting in the error metric $\sum_{i=1}^\numsourcept \Bigl(D_i(\mathbf{R}, \mathbf{t})\Bigr)^p$ that is used in the sparse ICP algorithm~\citemain{BouazizTP13}. 
Like classical ICP, the sparse ICP algorithm alternates between closest point query and alignment update. The alignment problem is similar to Eq.~\eqref{eq:Alignment} but is based on $\ell_p$ distance instead. The problem is solved using the alternating direction method of multipliers (ADMM) since there is no closed-form solution.
Although sparse ICP produces more accurate results, the use of ADMM incurs a much higher computational cost. Moreover, the ADMM solver requires  $d \cdot \numsourcept$ auxiliary variables and $d \cdot \numsourcept$ dual variables, which can significantly increase the memory footprint.

\begin{SCfigure}[0.8][!t]
	\centering
	\caption{The graphs of function $\psi_{\nu}(x)$ with different parameters. As $\nu$ decreases, the function $\psi_{\nu}$ approaches the $\ell_{0}$ norm.}
	\includegraphics[width=0.5\columnwidth]{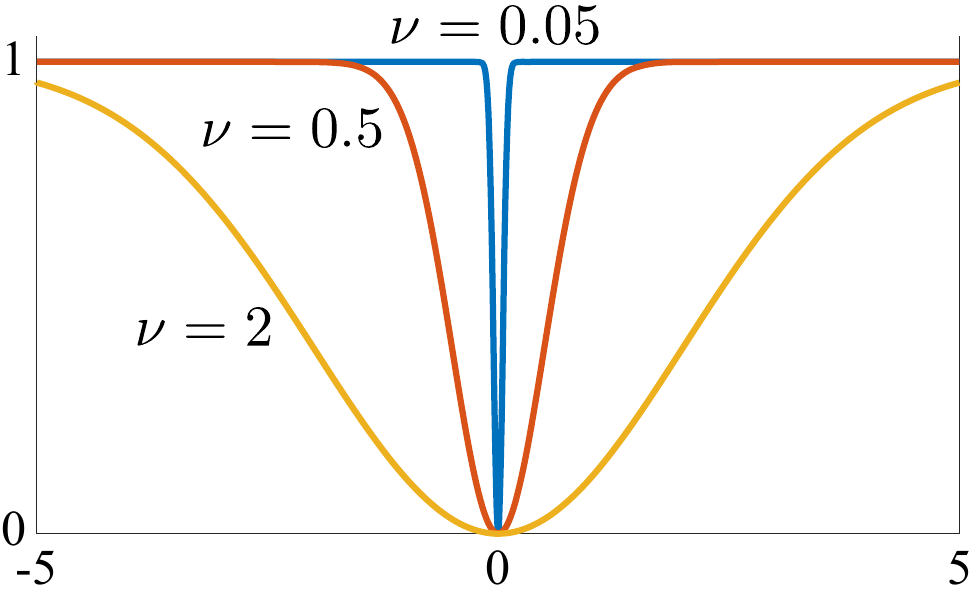}
	\label{fig:RegularizationFunction}
\end{SCfigure}

In this paper, we adopt a different robust error metric that does not incur high computational overhead. Specifically, we formulate the registration problem as
\begin{equation}
\min_{\mathbf{R}, \mathbf{t}} ~~ \sum\nolimits_{i=1}^\numsourcept \psi_{\nu}\bigl( D_i(\mathbf{R}, \mathbf{t}) \bigr) + I_{SO(d)} (\mathbf{R}),
\label{eq:WelschRegistration}
\end{equation}
where $\psi_{\nu}$ is the Welsch's function~\citemain{Holland77}:
\begin{equation}
\label{eq:welsch}
\psi_{\nu}(x) = 1 - \exp\Bigl(-\frac{x^{2}}{2 \nu^2}\Bigr),
\end{equation}
and $\nu > 0$ is a user-specified parameter.
Fig.~\ref{fig:RegularizationFunction} shows the graphs of $\psi_{\nu}$ with different values of $\nu$.
Since $\psi_{\nu}(x)$ is monotonically increasing on $[0, + \infty)$, our formulation penalizes deviation between the point sets. At the same time, $\psi_{\nu}$ is upper bounded by $1$, so our metric is not sensitive to large deviations caused by outliers and partial overlaps. Moreover, when $\nu$ approaches zero, $\sum_{i=1}^\numsourcept \psi_{\nu}\bigl( D_i(\mathbf{R}, \mathbf{t}) \bigr)$ approaches the $\ell_0$-norm of the vector $[D_1(\mathbf{R}, \mathbf{t}), \ldots, D_\numsourcept(\mathbf{R}, \mathbf{t})]$. Thus our formulation promotes sparsity of the point-wise distance between the point sets.
Recently, error metrics based on Welsch's function have been applied for robust filtering in image processing~\citemain{HamCP18} and geometry processing~\citemain{Zhang2019-SD}.

Although our formulation~\eqref{eq:WelschRegistration} is non-linear and non-convex, the problem can be solved using the same MM framework as classical ICP, by alternating between a correspondence step and an alignment step. The correspondence step is the same as classical ICP. In the alignment step, we utilize the closest points to construct the following surrogate for the target function~\eqref{eq:WelschRegistration} at the current transformation $(\mathbf{R}^{(k)}, \mathbf{t}^{(k)})$ (see Appendix~\ref{appx:SurrogateProof} for a proof):
\begin{equation}
	\sum\nolimits_{i=1}^\numsourcept \chi_{\nu} \Bigl(\|\mathbf{R} \sourcept_i + \mathbf{t} - \paringpt_i^{(k)}\| \Bigm| D_i(\mathbf{R}^{(k)}, \mathbf{t}^{(k)}) \Bigr) + I_{SO(d)} (\mathbf{R}),
	\label{eq:TargetSurrogateFunc}
\end{equation}
where $\chi_{\nu}(x \mid y)$ is a quadratic surrogate function for the Welsch's function at $y$ with the following form~\citemain{HamCP18}:
\begin{equation}
\chi_{\nu}(x \mid y)=\psi_{\nu}(y)+\frac{x^2-y^2}{2 \nu^2}\exp\Bigl(-\frac{y^{2}}{2 \nu^2}\Bigr).
\label{eq:WelschSurrogate}
\end{equation}
We minimize the surrogate function~\eqref{eq:TargetSurrogateFunc} to update the transformation, resulting in the following problem: 
\begin{equation}
\begin{aligned}
	&(\mathbf{R}^{(k+1)}, \mathbf{t}^{(k+1)}) \\
	=~ &\argmin_{\mathbf{R}, \mathbf{t}} \sum\nolimits_{i=1}^\numsourcept \omega_i \left\|\mathbf{R} \sourcept_i + \mathbf{t} - \paringpt_i^{(k)}\right\|^2 + I_{SO(d)} (\mathbf{R}),
\end{aligned}
\label{eq:WelschAlignment}	
\end{equation}
where $\omega_i = \exp\left(-\|\mathbf{R}^{(k)} \sourcept_i + \mathbf{t}^{(k)} - \paringpt_i^{(k)}\|^2/(2 \nu^2)\right)$.
The alignment step~\eqref{eq:WelschAlignment} minimizes a weighted sum of squared distance between the points $\{\sourcept_i\}$ and $\{\paringpt_i^{(k)}\}$. It can be solved in closed form via SVD~\citemain{RigidSVD}.
Similar to classical ICP, our MM solver decreases the target energy in each iteration and converges to a local minimum. Using the same approach as in Section~\ref{eq:AAForICP}, we improve its convergence rate by applying Anderson acceleration to the parameterization of rigid transformations in $\sea{d}$, using the same stabilization strategy that checks the target function value for the accelerated value.

\begin{figure}[!t]
	\centering
	\includegraphics[width=\columnwidth]{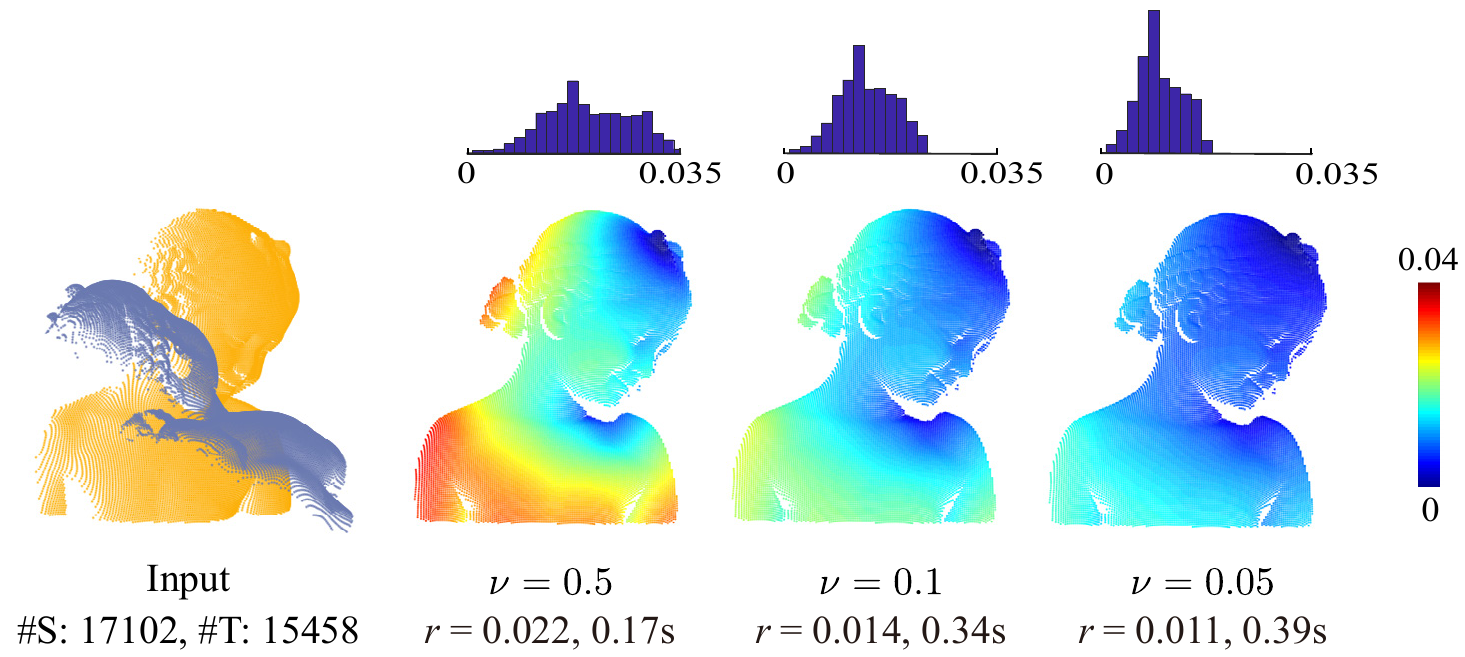}
	\caption{Registration results via optimization~\eqref{eq:WelschRegistration} with different values of parameter $\nu$, on a pair of point clouds with partial overlap. The color-coding shows the deviation between the transformed positions of each source point using the computed alignment and ground-truth alignment, with the histograms showing the distribution of the deviation among the source points. 
		For this model, a smaller value of $\nu$ leads to a more accurate result.}
	\label{fig:diffnu_pc}
\end{figure}

\begin{algorithm}[t]
	\KwIn{
		~ $\mathbf{T}^{(0)}$: initial transformation for $\sourceset$; \\
		~ $m$: the number of previous iterates used for Anderson acceleration; \\
		~ $\correspondence(\mathbf{T})$: computation of all closest points according via Eq.~\eqref{eq:CorresponcePoint} using transformation $\mathbf{T}$;\\
		~ $\alignment(\widehat{\mathbf{Q}}, \mathbf{T}, \nu)$: new transformation via Eq.~\eqref{eq:WelschAlignment} using current transformation $\mathbf{T}$ and closest points $\widehat{\mathbf{Q}}$;\\
		~ $E_{\nu}(\widehat{\mathbf{Q}}, \mathbf{T})$: target energy for transformation $\mathbf{T}$ and closest points $\widehat{\mathbf{Q}}$:
		$
		E_{\nu}(\widehat{\mathbf{Q}}, \mathbf{T}) = \sum\nolimits_{i=1}^M \psi_{\nu}( \|\mathbf{R}\sourcept_i + \mathbf{t} - \paringpt_i\| )
		$;\\
		~ $I_{\nu}, \epsilon_{\nu}$: maximum number of iterations and the convergence threshold of $\mathbf{T}$ for a given parameter $\nu$. 
	}
	\BlankLine
	$k = 1$; \quad $\nu = \nu_{max}$; \quad
	$\widehat{\mathbf{Q}}^{(0)} = \correspondence(\mathbf{T}^{(0)})$\;
	\While{\texttt{TRUE}}{
		$k_{\textrm{start}} = k - 1$; ~~
		$E_{\textrm{prev}} = +\infty$\;	
		$\mathbf{T'} = \alignment(\widehat{\mathbf{Q}}^{(k-1)}, \mathbf{T}^{(k-1)}, \nu)$\;
		$\mathbf{T}^{(k)} = \mathbf{T}'$;
		$~~\widehat{\mathbf{Q}}^{(k)} = \correspondence(\mathbf{T}^{(k)})$\;
		$G^{(k-1)} = \log(\mathbf{T'})$;
		$~~F^{(k-1)} = G^{(k-1)} - \log(\mathbf{T}^{(k-1)})$\;
		
		\While{$k - k_{\mathrm{start}} \leq I_{\nu}$}
		{			
			\tcp{Ensure $\mathbf{T}^{(k)}$ decreases the energy}
			\If{$E_{\nu}(\widehat{\mathbf{Q}}^{(k)}, \mathbf{T}^{(k)}) \geq E_{\mathrm{prev}}$}
			{
				$\mathbf{T}^{(k)} = \mathbf{T'}$;
				~~
				$\widehat{\mathbf{Q}}^{(k)} = \correspondence(\mathbf{T}^{(k)})$\;
			}
			$E_{\textrm{prev}} = E_{\nu}(\widehat{\mathbf{Q}}^{(k)}, \mathbf{T}^{(k)})$\;
			
			\BlankLine
			\tcp{Check convergence}
			$\mathbf{T'} = \alignment(\widehat{\mathbf{Q}}^{(k)}, \mathbf{T}^{(k)}, \nu)$\;
			\lIf{$\|\mathbf{T} - \mathbf{T'}\|_F < \epsilon_{\nu}$}
			{
				break
			}
			\tcp{Anderson acceleration}
			$G^{(k)} = \log(\mathbf{T'})$;~
			$F^{(k)} = G^{(k)} - \log(\mathbf{T}^{(k)})$\;
			$m_k = \min(k - k_{\textrm{start}}, m)$\;
			$(\theta_1^\ast, \ldots, \theta_{m_k}^\ast) = \argmin
			\| F^{(k)} - \sum\nolimits_{j=1}^{m_k} \theta_j (F^{(k-j+1)} - F^{(k-j)}) \|_F^2$\;
			$\mathbf{T}^{(k+1)}= \exp\bigl(G^{(k)} - \sum_{j=1}^m \theta_j^\ast (G^{(k-j+1)} - G^{(k-j)}) \bigr)$\;
			
			
			$\widehat{\mathbf{Q}}^{(k+1)} = \correspondence(\mathbf{T}^{(k+1)})$\;
			
			
			$k = k+1$\;
		}
		
		\lIf{$\nu == \nu_{\min}$}
		{
			\Return{$\mathbf{T}^{(k)}$}	
		}
		
		$\nu=\max(\nu/2, \nu_{\min})$;$~~~~k=k+1$;
		
	}
	\caption{Robust point-to-point ICP using Welsch's function and Anderson acceleration.}
	\label{alg:AA-Robust-ICP}
\end{algorithm}
Our approach has a similar structure as the iteratively reweighted least squares (IRLS) method that minimizes the $\ell_p$-norm ($p < 1$) for compressive sensing~\citemain{Chartrand2008}. Similar to IRLS, we solve a weighted least squares problem, with the weights $\omega_i$ for a point $\sourcept_i$ updated in each iteration according to its current distance to the corresponding point. Since the weight is a Gaussian function with variance $\nu^2$, a point $\sourcept_i$ with larger distance from the target point set receives a lower weight. Moreover, according to the well-know three-sigma rule, when the distance is larger than $3 \nu$, the weight $\omega_i$ is small enough such that the term for $\sourcept_i$ has little influence to the target function and $\sourcept_i$ is effectively excluded from the current alignment problem. In this way, the optimization allows some source points to be far away from the target point set, and is robust to outliers and partial overlaps.

Some ICP variants improve robustness by excluding from the alignment step the point pairs with large deviation between their positions or normals~\citemain{RusinkiewiczL01}. It was observed in~\citemain{BouazizTP13} that such methods can be difficult to tune or increase the number of local minima. Our method also excludes point pairs with large positional difference, but using a Gaussian weight that gradually decreases as the point pair becomes further apart. It can be considered as a soft thresholding approach that weakly penalizes outliers, which can lead to more stable results~\citemain{BouazizTP13}. Indeed, we observe in experiments that our robust methods and sparse ICP tend to produce more accurate results than the symmetric ICP method~\citemain{Rusinkiewicz19} which is based on outlier rejection; see Section~\ref{sec:results} for details.

Compared to $\ell_p$-norm minimization ($0 < p < 1$), our formulation and solver also offer benefits in stability and convergence guarantee. For our weighted least-squares problem~\eqref{eq:WelschAlignment}, all the Gaussian weights $\{\omega_i\}$ have values within the range $(0, 1]$. In contrast, an IRLS solver for $\ell_p$-norm minimization would assign a weight $\|\mathbf{R}^{(k)} \sourcept_i + \mathbf{t}^{(k)} - \paringpt_i^{(k)}\|^{p-2}$ to the point $\mathbf{p}_i$, which could go to infinity and cause instability when the alignment error for $\mathbf{p}_i$ approaches zero~\citemain{BouazizTP13}. According to~\citemain{BouazizTP13}, sparse ICP performs $\ell_p$-norm minimization using ADMM instead of IRLS because of concern about such instability.
In addition, the convergence of IRLS and ADMM for non-convex $\ell_p$-norm minimization requires strong assumptions about the problem such as the Kurdyka-{\L}ojasiewicz property~\citemain{Ochs2015,wang2019global}, whereas our MM solver is guaranteed to converge.

For our algorithm, the parameter $\nu$ plays an important role in achieving good performance. A smaller $\nu$ helps to attenuate the influence from outliers and partial overlaps (e.g., see Fig.~\ref{fig:diffnu_pc}). On the other hand, a larger $\nu$ in the initial stage helps to include more point pairs in the alignment step and avoid undesirable local minima. Therefore, we gradually decrease $\nu$ during the iterations, so that the algorithm first performs more global alignment with a larger number of pairs, and then reduces the influence from the pairs with large deviation to achieve robust alignment. Specifically, we choose two values $\nu_{\max}$ and $\nu_{\min}$ as the upper and lower bounds of $\nu$. We start by setting $\nu = \nu_{\max}$ and running our MM algorithm until the change in the transformation matrix $\mathbf{T}$ is smaller than a threshold ($10^{-5}$ by default) or the iteration count exceeds an upper limit ($1000$ by default). Then we decrease the value of $\nu$ by half, and run the MM algorithm again until the same termination criterion is met. The process is repeated until the lower bound $\nu_{\min}$ is reached. Algorithm~\ref{alg:AA-Robust-ICP} summarizes our method with a decreasing $\nu$.

\begin{figure}[!t]
	\centering
	\includegraphics[width=0.85\columnwidth]{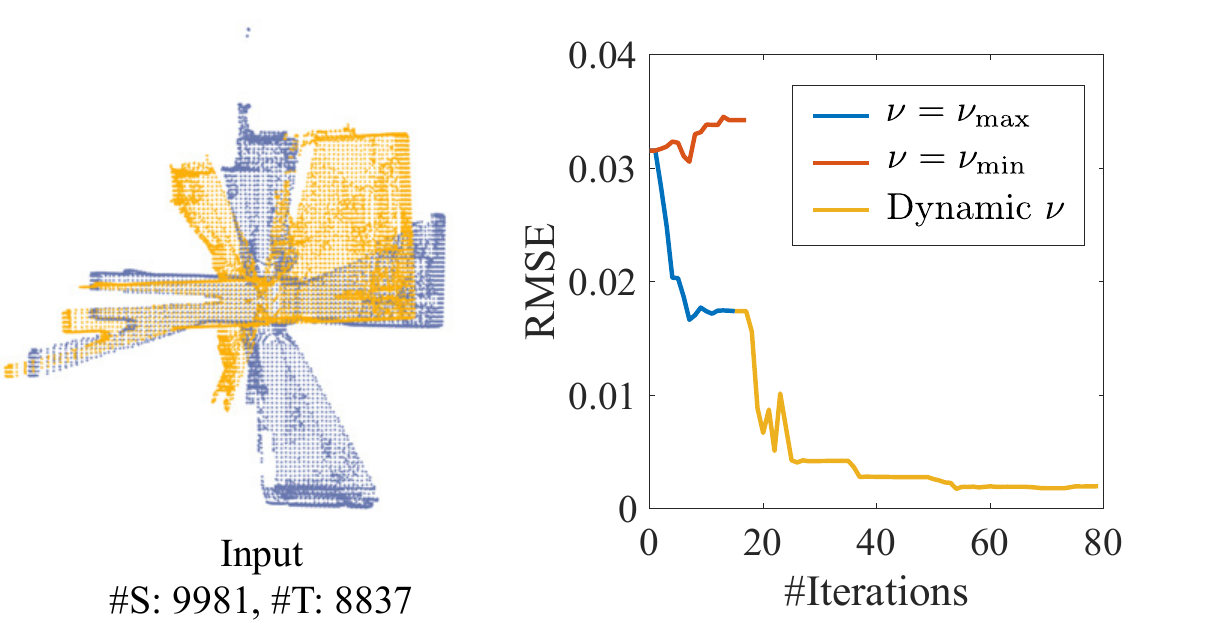}
	\caption{RMSE plots for optimization~\eqref{eq:WelschRegistration} with different settings of $\nu$, on a pair of point clouds in the `Apartment' sequence from the ETH laser registration dataset~\protect\citemain{Pomerleau2012}. Gradually reducing $\nu$ from $\nu_{\max}$ to $\nu_{\min}$ results in the lowest RMSE.}
	\label{fig:DynamicNu}
\end{figure}

To choose $\nu_{\max}$, we compute the median $\overline{D}^{(0)}$ among all initial point-wise distance $\{D_i(\mathbf{R}^{(0)}, \mathbf{t}^{(0)})\}$, and set $\nu_{\max} = 3 \cdot \overline{D}^{(0)}$.
In our experiments, this setting makes $\nu_{\max}$ large enough to include most point pairs into the alignment process except for outliers with significant deviation.
For $\nu_{\min}$, we note that the two point sets may sample the same surface at different locations, and $\nu_{\min}$ should be large enough to accommodate the deviation due to sampling. Therefore, we first compute the median distance from each point $\targetpt_i\in \targetset$ to its six nearest points on $\targetset$, and take the median $\overline{E}_{\targetset}$ of all such median values. Then we set $\nu_{\min} = {\overline{E}_{\targetset}}/{3 \sqrt{3}}$ (see Appendix~\ref{appx:SettingNu} for the rationale).

Fig.~\ref{fig:DynamicNu} illustrates the effectiveness of our $\nu$-update strategy, by comparing its RMSE plot with those resulting from a fixed parameter $\nu = \nu_{\max}$ and $\nu = \nu_{\min}$, respectively. Here a fixed $\nu = \nu_{\min}$ results in a large RMSE, because such a small $\nu$ will lead to a small weight for most point pairs, effectively excluding them from the alignment step and producing an erroneous result. Fixing $\nu = \nu_{\max}$ can reduce the final RMSE as it includes more points into the alignment; however, it fails to exclude some outliers so the RMSE is still large. A decreasing $\nu$ gradually removes outliers from the alignment process, resulting in a much smaller RMSE.

\section{Extension to Point-to-Plane ICP}
\label{sec:PointToPlane}

The classical ICP algorithm discussed in Section~\ref{sec:AA} is often called the ``point-to-point'' ICP, since its alignment step minimizes the distance from the source points to their corresponding target points. Another popular ICP variant in $\mathbb{R}^3$, often called the ``point-to-plane'' ICP, minimizes the distance from the source points to the tangent planes at the target points instead in the alignment step~\citemain{Chen1992}:
\begin{equation}
\hspace*{-1em}
\begin{aligned}
&(\mathbf{R}^{(k+1)}, \mathbf{t}^{(k+1)}) \\
= &\argmin_{\mathbf{R}, \mathbf{t}} \sum_{i=1}^\numsourcept \left((\mathbf{R} \sourcept_i + \mathbf{t} - \paringpt_i^{(k)}) \cdot \paringnormal_i^{(k)}\right)^2 + I_{SO(3)} (\mathbf{R}),
\end{aligned}
\label{eq:PointToPlaneAlignment}
\end{equation}
where $\paringnormal_i^{(k)}$ is the normal at $\paringpt_i^{(k)}$ for the underlying surface of the target point set. 
Point-to-plane ICP can be considered as solving an optimization problem
\begin{equation}
\min_{\mathbf{R}, \mathbf{t}} \sum\nolimits_{i=1}^\numsourcept \left(H_i(\mathbf{R}, \mathbf{t})\right)^2 + I_{SO(3)} (\mathbf{R}),
\label{eq:PointToPlaneOptimization}
\end{equation}
where $H_i(\mathbf{R}, \mathbf{t})$ is the signed distance from the point $\mathbf{R} \mathbf{p}_i + \mathbf{t}$ to the tangent plane at its closest point in $\targetset$.
Since the tangent plane provides a local linear approximation of the underlying surface, point-to-plane ICP can achieve faster convergence~\citemain{PottmannHYH06}. On the other hand, it suffers from the same issue of robustness to outliers and partial overlaps. Similar to Section~\ref{sec:AA}, we can improve its robustness by adopting a robust metric based on Welsch's function $\psi_{\nu}$:
\begin{equation}
\min_{\mathbf{R}, \mathbf{t}} \sum\nolimits_{i=1}^\numsourcept \psi_{\nu} \left(H_i(\mathbf{R}, \mathbf{t})\right) + I_{SO(3)} (\mathbf{R}).
\label{eq:WelschPointToPlaneOptimization}
\end{equation}
This is solved by alternating between a correspondence step the same as point-to-point ICP, and an assignment step that solves the following problem:
\begin{equation}
\min_{\mathbf{R}, \mathbf{t}} \sum\nolimits_{i=1}^\numsourcept \psi_{\nu}\left((\mathbf{R} \sourcept_i + \mathbf{t} - \paringpt_i^{(k)}) \cdot \paringnormal_i^{(k)}\right) + I_{SO(3)} (\mathbf{R}).
\label{eq:WelschPointToPlaneAlignment}
\end{equation}
Similar to Section~\ref{sec:AA}, we replace the target function above with a surrogate function to derive a proxy problem:
\begin{equation}
	\min_{\mathbf{R}, \mathbf{t}} \sum\nolimits_{i=1}^\numsourcept  \gamma_i \left((\mathbf{R} \sourcept_i + \mathbf{t} - \paringpt_i^{(k)}) \cdot \paringnormal_i^{(k)}\right)^2 + I_{SO(3)} (\mathbf{R}),
	\label{eq:ProxyPointToPlaneAlignment}
\end{equation}
where $\gamma_i = \exp\left(-((\mathbf{R}^{(k)} \sourcept_i + \mathbf{t}^{(k)} - \paringpt_i^{(k)}) \cdot \paringnormal_i^{(k)})^2/(2 \nu^2)\right)$.
There is no closed-form solution to this problem. So we rewrite it as an optimization for the $\sea{3}$ parameterization:
\begin{equation}
	\min_{\liealgparam{\mathbf{T}}} \sum\nolimits_{i=1}^\numsourcept  \gamma_i \left(B_i^{(k)}(\liealgparam{\mathbf{T}})\right)^2.
	\label{eq:LieAlgebraPointToPlaneAlignment}
\end{equation}
Here $\liealgparam{\mathbf{T}} \in \mathbb{R}^6$ denotes the actual variables for the $\sea{3}$ element $\liealg{\mathbf{T}}$ in Eq.~\eqref{eq:Matrixsed} (three variables for each of the submatrices $\mathbf{S}$ and $\mathbf{u}$, respectively), and $B_i^{(k)}(\liealgparam{\mathbf{T}})$ is the signed distance from $\mathbf{R} \sourcept_i + \mathbf{t}$ to the tangent plane at $\paringpt_i^{(k)}$. We then linearize $B_i^{(k)}$ using its first-order Taylor expansion
\begin{equation}
	{B}_i^{(k)}(\liealgparam{\mathbf{T}}) \approx {B}_i^{(k)}(\liealgparam{\mathbf{T}}^{(k)}) + (\mathbf{J}_i^{(k)})^T (\liealgparam{\mathbf{T}} - \liealgparam{\mathbf{T}}^{(k)}),
	\label{eq:Linearization}
\end{equation}
where $\liealgparam{\mathbf{T}}^{(k)}$ is the $\sea{3}$ variable for $({\mathbf{R}}^{(k)}, \mathbf{t}^{(k)})$, and $\mathbf{J}_i^{(k)}$ is the gradient of ${B}_i^{(k)}$ at $\liealgparam{\mathbf{T}}^{(k)}$ (see Appendix~\ref{appx:Gradient} for its calculation). Substituting the linearization into Eq.~\eqref{eq:LieAlgebraPointToPlaneAlignment}, we obtain a quadratic problem that reduces to a linear system
\begin{equation}
	\begin{aligned}
	&\left(\sum\nolimits_{i=1}^\numsourcept \gamma_i \mathbf{J}_i^{(k)} (\mathbf{J}_i^{(k)})^T \right) \liealgparam{\mathbf{T}}\\
	= & 
	~\sum\nolimits_{i=1}^\numsourcept \gamma_i \mathbf{J}_i^{(k)} \left( {B}_i^{(k)}(\liealgparam{\mathbf{T}}^{(k)})  - 
	(\mathbf{J}_i^{(k)})^T \liealgparam{\mathbf{T}}^{(k)}
	\right).
	\end{aligned}
	\label{eq:GNSystem}
\end{equation}
The solution $\liealgparam{\mathbf{T}}_{\ast}^{(k)}$ to this system will be taken as a candidate for the updated transformation. Due to the linearization, $\liealgparam{\mathbf{T}}_{\ast}^{(k)}$ may increase the target function~\eqref{eq:WelschPointToPlaneOptimization}. Therefore, we perform line search along the direction $\liealgparam{\mathbf{T}}_{\ast}^{(k)} - \liealgparam{\mathbf{T}}^{(k)}$ to find a new transformation that decreases the target function. If such a transformation cannot be found after the maximum number of line-search steps is reached, then the step size with the lowest target function value will be used.

\begin{algorithm}[t]
		\KwIn{
			~ $\liealgparam{\mathbf{T}}^{(0)}$: initial transformation parameters; \\
			~ $m$:  the number of previous iterates used for Anderson acceleration; \\
			~ $\pplmap(\cdot)$: the mapping defined in Eq.~\eqref{eq:GNMapping};\\
			~ $\widetilde{E}_{\nu}(\liealgparam{\mathbf{T}})$: target energy of problem~\eqref{eq:WelschPointToPlaneOptimization} for parameters $\liealgparam{\mathbf{T}}$;\\
			~ $l_{\max}$: maximum number of inner line search steps;\\
			~ $I_{\nu}, \epsilon_{\nu}$: maximum number of iterations and the convergence threshold for a given parameter $\nu$. 
		}
		$k = 1$; \quad $\nu = \nu_{max}$\;
		\While{\texttt{TRUE}}{
			$k_{\textrm{start}} = k - 1$; ~
			$E_{\textrm{prev}} = +\infty$;~
			$\liealgparam{\mathbf{T}}_{\ast}^{(k)} = \pplmap(\liealgparam{\mathbf{T}}^{(k-1)})$ \;
			$G^{(k-1)} = \liealgparam{\mathbf{T}}^{(k)} = \liealgparam{\mathbf{T}}_{\ast}^{(k)}$;
			$~F^{(k-1)} = G^{(k-1)} - \liealgparam{\mathbf{T}}^{(k-1)}$\;
			
			\While{$k - k_{\mathrm{start}} \leq I_{\nu}$}
			{			
				\tcp{Check energy decrease}
				$E = \widetilde{E}_{\nu}(\liealgparam{\mathbf{T}}^{(k)})$\;
				\If{$E \geq E_{\mathrm{prev}}$}
				{
					\tcp{Perform line search}
					$\tau = 1$;~~ $l = 1$\;
					\While{$l \leq l_{\max}$}
					{
						$\liealgparam{\mathbf{T}}_{\textrm{trial}} = \liealgparam{\mathbf{T}}^{(k-1)} + \tau (\liealgparam{\mathbf{T}}_{\ast}^{(k)} - \liealgparam{\mathbf{T}}^{(k-1)})$\;
						$E_{\textrm{trial}} = \widetilde{E}_{\nu}(\liealgparam{\mathbf{T}}_{\textrm{trial}})$\;
						\If{$E_{\mathrm{trial}} < E$}
						{
							$E = E_{\mathrm{trial}}$;~ $\liealgparam{\mathbf{T}}^{(k)} = \liealgparam{\mathbf{T}}_{\textrm{trial}}$\;
						}
						\lIf{$E_{\mathrm{trial}} < E_{\mathrm{prev}}$}
						{
							break
						}
					}
				}
				$E_{\textrm{prev}} = E$\;
				
				\BlankLine
				\tcp{Check convergence}
				$\liealgparam{\mathbf{T}}_{\ast}^{(k+1)} = \pplmap(\liealgparam{\mathbf{T}}^{(k)})$ \;
				\lIf{$\|\liealgparam{\mathbf{T}}_{\ast}^{(k+1)} - \liealgparam{\mathbf{T}}^{(k)}\| < \epsilon_{\nu}$}
				{
					break;
				}
				
				\tcp{Anderson acceleration}
				$G^{(k)} = \liealgparam{\mathbf{T}}_{\ast}^{(k+1)}$;~
				$F^{(k)} = G^{(k)} - \liealgparam{\mathbf{T}}^{(k)}$\;
				$m_k = \min(k - k_{\textrm{start}}, m)$\;
				$(\theta_1^\ast, \ldots, \theta_{m_k}^\ast) = \argmin
				\| F^{(k)} - \sum\nolimits_{j=1}^{m_k} \theta_j (F^{(k-j+1)} - F^{(k-j)}) \|^2$\;
				$\liealgparam{\mathbf{T}}^{(k+1)}= \exp\bigl(G^{(k)} - \sum_{j=1}^m \theta_j^\ast (G^{(k-j+1)} - G^{(k-j)}) \bigr)$\;

				$k = k+1$\;
			}
			
			\lIf{$\nu == \nu_{\min}$}
			{
				\Return{$\liealgparam{\mathbf{T}}^{(k)}$}	
			}
			
			$\nu=\max(\nu/2, \nu_{\min})$;$~~~~k=k+1$;
	}
	\caption{Robust point-to-plane ICP using Welsch's function and Anderson acceleration.}
	\label{alg:AA-PointToPlane-ICP}
\end{algorithm}

Similar to Section~\ref{sec:AA}, we apply Anderson acceleration to speed up the convergence. We note that the mapping from the current variable $\liealgparam{\mathbf{T}}^{(k)}$ to the candidate update $\liealgparam{\mathbf{T}}_{\ast}^{(k)}$, which amounts to finding the closest points $\{\paringpt_i^{(k)}\}$ according to $\liealgparam{\mathbf{T}}^{(k)}$ and solving the linear system~\eqref{eq:GNSystem}, can be written as
\begin{equation}
	\liealgparam{\mathbf{T}}_{\ast}^{(k)} = \pplmap(\liealgparam{\mathbf{T}}^{(k)}).
	\label{eq:GNMapping}
\end{equation}
Then for a local minimum $(\mathbf{R}^{\ast}, \mathbf{t}^{\ast})$ of the target function~\eqref{eq:WelschPointToPlaneOptimization}, the corresponding $\sea{3}$ variable $\liealgparam{\mathbf{T}}^{\ast}$ should be a fixed point of $\pplmap{}$. Therefore, we apply Anderson acceleration to $\liealgparam{\mathbf{T}}^{(k-m)}, \ldots, \liealgparam{\mathbf{T}}^{(k)}$ and $\pplmap(\liealgparam{\mathbf{T}}^{(k-m)}), \ldots, \pplmap(\liealgparam{\mathbf{T}}^{(k)})$ to obtain an accelerated value $\liealgparam{\mathbf{T}}_{\textrm{AA}}$. If $\liealgparam{\mathbf{T}}_{\textrm{AA}}$ decreases the target function~\eqref{eq:WelschPointToPlaneOptimization}, then we accept it as the new iterate $\liealgparam{\mathbf{T}}^{(k+1)}$. Otherwise, we perform line search as described previously. 
Algorithm~\ref{alg:AA-PointToPlane-ICP} summarizes our robust point-to-plane ICP solver.
Similarly to Algorithm~\ref{alg:AA-Robust-ICP}, we start with $\nu = \nu_{\max}$ and gradually decreases it until the lower bound $\nu_{\min}$ is reached. 
For each given $\nu$ value, the solver is run until the change in the transformation matrix is smaller than a threshold ($10^{-5}$ by default) or the iteration count reaches a limit ($6$ for $\nu_{\max}$, then incremented by $1$ each time $\nu$ is changed, but no larger than $10$).
We set $\nu_{\max}$ to be three times the median distance from the source points to the tangent planes at their corresponding points in the initial iteration. 
To determine $\nu_{\min}$, we first compute for each point $\targetpt \in \targetset$ the median distance from its six nearest neighbors in $\targetset$ to its tangent plane; then we take the median $\overline{H}_{\targetset}$ of all such values and  set $\nu_{\min}  = \overline{H}_{\targetset} / 6$ (see Appendix~\ref{appx:SettingNu} for the rationale).

\section{Results}
\label{sec:results}

In this section, we compare the performance of our methods with existing ICP-based methods including AA-ICP~\citemain{podto-aaicp-2017pre}, sparse ICP~\citemain{BouazizTP13}, and symmetric ICP~\citemain{Rusinkiewicz19}.
Our comparison includes both point-to-point and point-to-plane ICP methods and their variants. In the following, we will denote the point-to-point ICP and its variants as ``ICP'', whereas point-to-plane ICP and its variants will be denoted as ``ICP-\emph{l}''.
For sparse ICP and symmetric ICP, we use the source codes released by the authors\footnote{\url{https://github.com/OpenGP/sparseicp}}
\footnote{\url{https://gfx.cs.princeton.edu/pubs/Rusinkiewicz_2019_ASO/icptests-1.0.zip}}.
For symmetric ICP, we use the formulation that does not rotate the normals ($\mathcal{E}_{symm}$ as defined in the paper).
Besides ICP-based methods, we also compare with other methods including CPD~\citemain{Myronenko2010} and GMM-Reg~\citemain{Jian2011} based on statistical frameworks, Teaser++~\citemain{Yang2020} which uses truncated least squares optimization, as well as DCP~\citemain{Wang2019} and DGR~\citemain{Choy2020} based on deep learning, using their open-source implementations
\footnote{\url{https://github.com/gadomski/cpd}}
\footnote{\url{https://github.com/bing-jian/gmmreg}}
\footnote{\url{https://github.com/MIT-SPARK/TEASER-plusplus}}
\footnote{\url{https://github.com/WangYueFt/dcp}}
\footnote{\url{https://github.com/chrischoy/DeepGlobalRegistration}}.
We implement our methods in C++, using the \eigen{} library~\citemain{eigenweb} for linear algebra operations. For each test problem, we normalize the input data by aligning the centroids of the point clouds and uniformly scaling them such that their bounding box has diagonal length 1. Unless stated otherwise, the point clouds are pre-aligned using Super4PCS~\citemain{Mellado2014} before the alignment is refined using different methods.
We test the methods on both synthetic and real-world datasets. 
For problems where the ground-truth alignment is known, we evaluate the registration accuracy using the RMSE value in Eq.~\eqref{eq-mse}.
Some methods (point-to-plane ICP and its variants, as well as symmetric ICP) require normals at the points. For synthetic data where the underlying surface is known, we use the surface normals as the normals at the points. Otherwise, we use the Point Cloud Library\footnote{https://pointclouds.org/} to estimate the normals using 30 nearest neighbors.  
The source codes for our methods are available at \url{https://github.com/yaoyx689/Fast-Robust-ICP}.
The two deep learning-based methods are run on a PC with a 20-core CPU at 3.3GHz, an NVIDIA RTX 2080 Ti and 128GB of RAM, whereas all other methods are run on a PC with a 6-core CPU at 3.6GHz and 16GB of RAM.
Detailed settings for each method are provided in Appendix~\ref{appx:Settings}.

\begin{figure*}[!t]
	\centering
	\includegraphics[width=\textwidth]{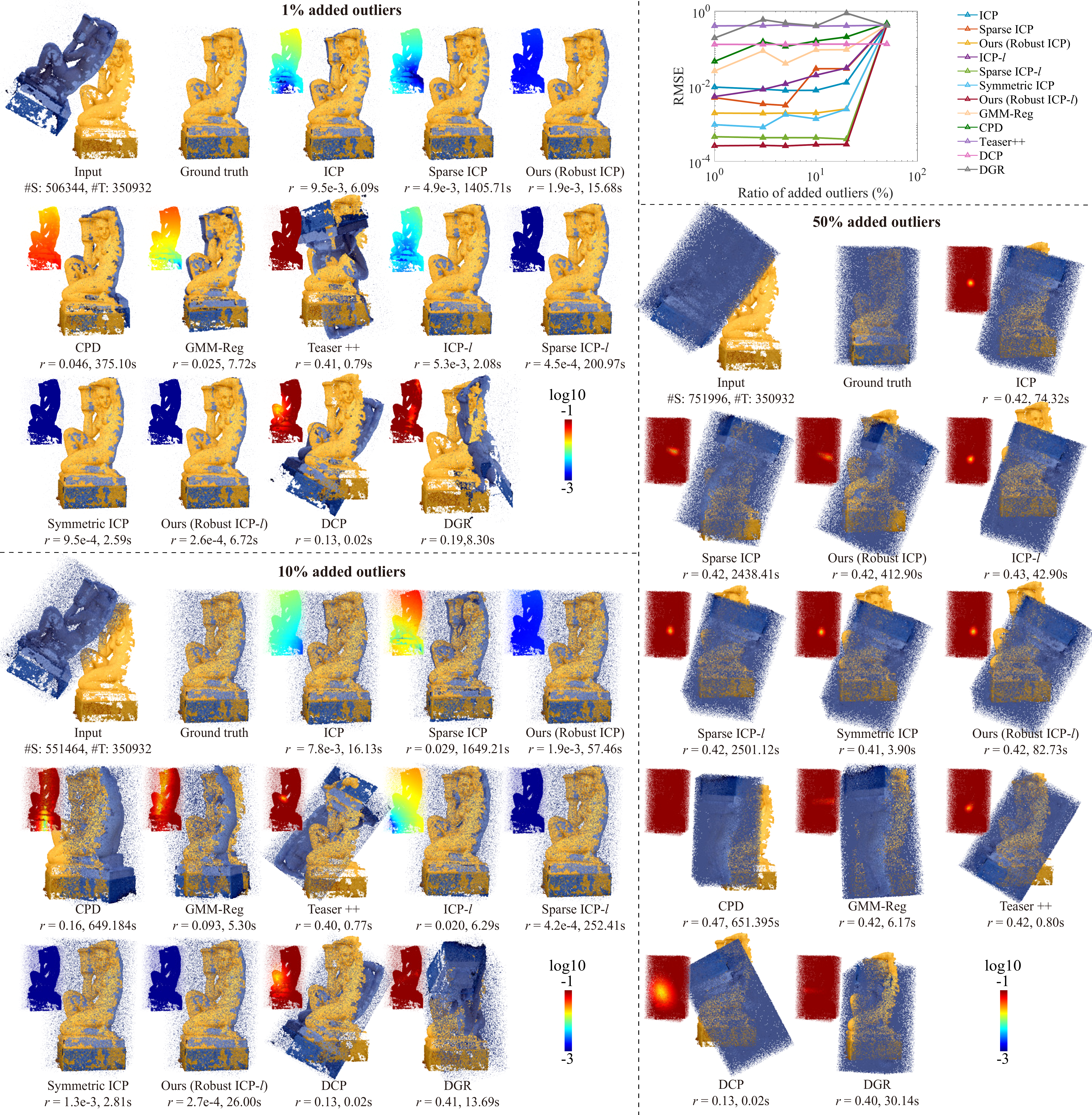}
	\caption{Comparison between different registration methods on partially overlapping point clouds with added noises and outliers, constructed using the Aquarius model from the EPFL statue dataset~\protect\citemain{EPFL2012}. The plot on the top-right shows the resulting RMSE values with different ratios ($1\%$, $3\%$, $5\%$, $20\%$ and $50\%$) of outliers added to the source point cloud.}
	\label{fig:aquarius_noise}
\end{figure*}

\subsection{Synthetic Data}

In Fig.~\ref{fig:teaser-monkeys}, we perform registration on two point sets with a small overlap, constructed using the monkey model from the EPFL statue dataset~\citemain{EPFL2012}. From the full model, we take the first $60\%$ of the points to create the source set, and the last $47\%$ with a random rigid transformation to construct the target set.
Our robust point-to-point and point-to-plane ICP methods achieve the lowest RMSE values among all methods, while being significantly faster than point-to-point ICP, the method that achieves the next lowest RMSE. In addition, our fast ICP achieves the same RMSE as classical ICP and AA-ICP with less computational time. Symmetric ICP also achieves better accuracy than point-to-point and point-to-plane ICP and their accelerated versions; it is faster than sparse ICP and our robust methods  but with worse accuracy. The saving in computational time from symmetric ICP is partly because its implementation only samples 200 pairs of valid corresponding points for the alignment step, which reduces the computational cost for large point clouds.

\begin{figure*}[!t]
	\centering
	\includegraphics[width=\textwidth]{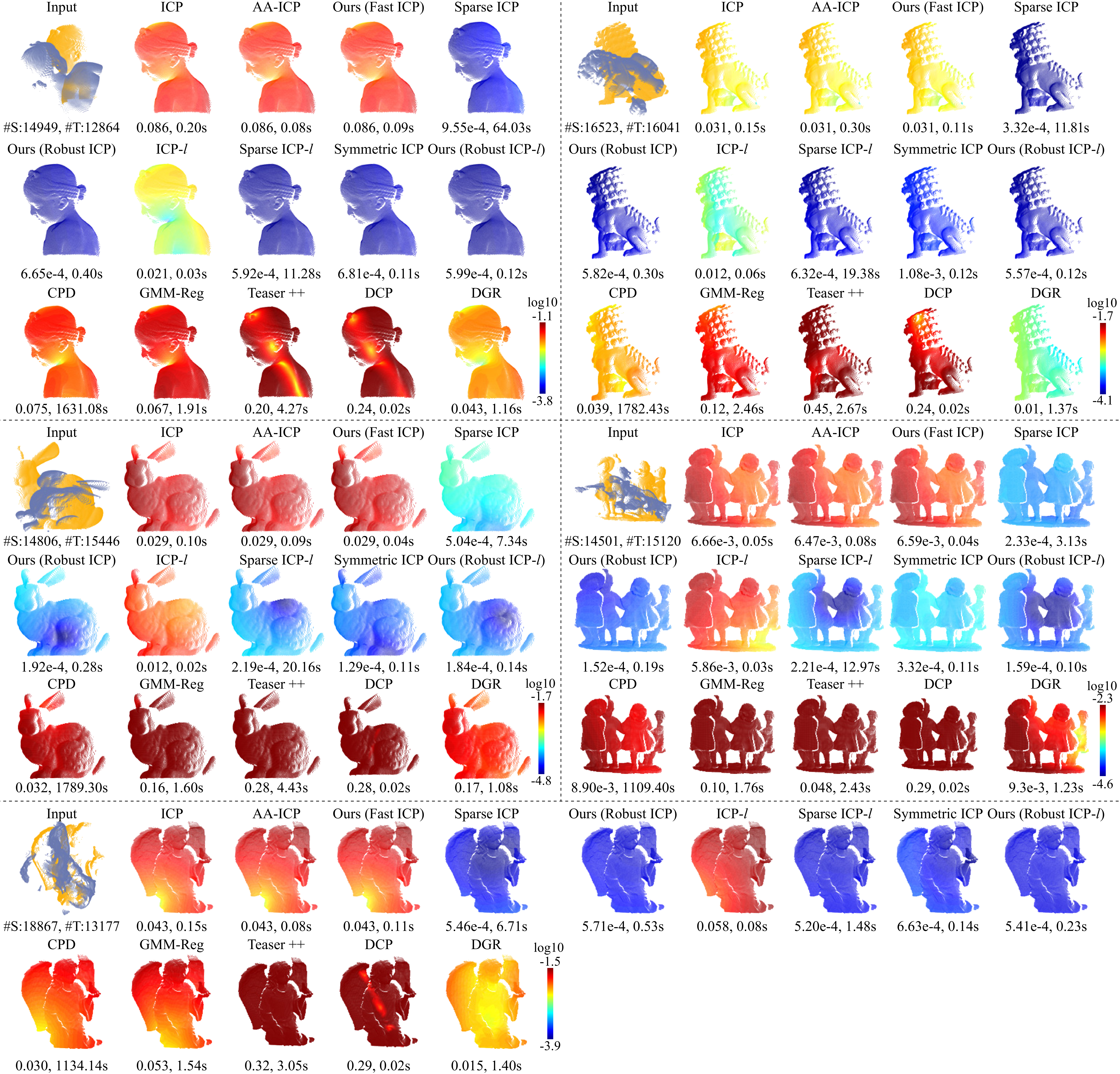}
	\caption{Examples of registration results using different methods on partially overlapping point clouds, with RMSE and computational time shown below each result. The log-scale color-coding visualizes the deviation between the computed alignment and the ground-truth alignment.}
	\label{fig:FGR_pc}
\end{figure*}

In Fig.~\ref{fig:aquarius_noise}, we test the methods on point sets that contain noises and outliers, which are constructed using the Aquarius model from the EPFL statue dataset.
Starting from the clean point cloud of the full model, 
we take the last $42\%$ of the points as the target point cloud, add Gaussian noises along their normal directions with the standard deviation being the average value of all points' median distance to their six nearest neighbors, and apply a random transformation. 
For the source point cloud, we take the first $60\%$ of the points from the full model, and add Gaussian noises in the same way as the target point cloud. To emulate outliers, we add $\mu \cdot \overline{M}$ random points to the source point cloud using a uniform distribution within its bounding box, where $\overline{M}$ is the number of source points before the addition, and $\mu$ is chosen to be $1\%$, $3\%$, $5\%$, $20\%$ and $50\%$, respectively. For problems with up to $\mu = 20\%$ added outliers, our robust point-to-point and point-to-plane methods outperform other point-to-point and point-to-plane ICP variants respectively in accuracy, with our robust point-to-plane method achieving the best accuracy among all methods. For $50\%$ added outliers, all methods result in similar RMSE values that indicate large registration error. For ICP-based methods, this is partly due to poor initial alignment produced by Super4PCS in the presence of outliers. With better initialization, our methods can still produce reasonable registration results (see Section~\ref{sec:limitations} for details).
It is also worth noting that Teaser++, which is aimed at problems with a large amount of outliers, performs poorly in this example. This is potentially because Teaser++ assumes a generative model where the deviation between corresponding points is due to a bounded noise, which is not obeyed by the randomly generated outliers here.

\begin{table*}[!t]
	\caption{Average computational time (in seconds) and average/median RMSE ($\times 10^{-3}$) for different registration methods on partially overlapping point cloud pairs constructed from five models, with five pairs for each model (see Fig.~\ref{fig:FGR_pc}). Best performance numbers are highlighted in bold fonts.}
	\label{Tab:FRG_Time_MSE}
	\setlength{\tabcolsep}{1.8pt}
	\centering
	{
		\fontsize{8}{9.6}\selectfont
		\begin{tabular}{ | c |  c  c | c  c | c  c | c  c  |  c c | c c | }
			\Xhline{1pt}
			\multicolumn{1}{ | c |}{\multirow{2}{*}{Dataset}}
			&\multicolumn{2}{ c |}{Bimba}	
			&\multicolumn{2}{ c | }{Children}
			&\multicolumn{2}{ c | }{Dragon}
			&\multicolumn{2}{ c |}{Angle}	
			&\multicolumn{2}{ c | }{Bunny}\\\cline{2-11}
			\multicolumn{1}{ | c |}{}
			&Time  	&RMSE	
			&Time 	 &RMSE	
			&Time  	&RMSE	
			&Time  	&RMSE
			&Time 	 &RMSE \\\hline
			ICP & \num{0.33} & \num{68}/\num{60}   & \num{0.12} & \num{9.8}/\num{6.7}   & \num{0.18} & \num{21}/\num{19}   & \num{0.12} & \num{13}/\num{5.6}   & \num{0.11} & \num{26}/\num{28}  \\ 
			AA-ICP & \num{0.13} & \num{68}/\num{60}   & \num{0.10} & \num{9.8}/\num{6.5}   & \num{0.16} & \num{21}/\num{19}   & \num{0.08} & \num{13}/\num{5.6}   & \num{0.10} & \num{26}/\num{28}  \\ 
			Ours (Fast ICP) & \num{0.12} & \num{68}/\num{60}   & \num{0.07} & \num{9.8}/\num{6.6}   & \num{0.12} & \num{21}/\num{19}   & \num{0.11} & \num{13}/\num{5.6}   & \num{0.06} & \num{26}/\num{28}  \\ 
			Sparse ICP & \num{37.90} & \num{67}/\num{27}   & \num{8.59} & \num{0.96}/\num{0.81}   & \num{24.42} & \textbf{\num{0.92}}/\num{0.95}   & \num{15.45} & \num{0.83}/\textbf{\num{0.97}}   & \num{24.06} & \num{0.94}/\num{0.71}  \\ 
			Ours (Robust ICP) & \num{0.96} & \textbf{\num{0.87}}/\num{0.67}   & \num{0.26} & \num{0.89}/\textbf{\num{0.62}}   & \num{0.27} & \num{0.93}/\textbf{\num{0.92}}   & \num{0.27} & \num{0.83}/\num{0.98}   & \num{0.34} & \num{0.85}/\num{0.69}  \\ 
			ICP-\emph{l} & \num{0.74} & \num{78}/\num{20}   & \num{0.07} & \num{16}/\num{5.9}   & \num{1.06} & \num{14}/\num{11}   & \num{0.07} & \num{13}/\num{2.5}   & \num{0.06} & \num{13}/\num{12}  \\ 
			Sparse ICP-\emph{l} & \num{20.97} & \num{3.4}/\textbf{\num{0.59}}   & \num{20.69} & \num{0.92}/\num{0.67}   & \num{14.25} & \num{0.96}/\num{0.94}   & \num{6.23} & \textbf{\num{0.82}}/\num{0.98}   & \num{10.68} & \num{0.81}/\textbf{\num{0.56}}  \\ 
			Symmetric ICP & \num{0.25} & \num{34}/\num{0.6}   & \num{0.19} & \textbf{\num{0.88}}/\num{0.64}   & \num{0.20} & \num{0.92}/\num{0.93}   & \num{0.20} & \num{0.82}/\num{0.98}   & \num{0.18} & \textbf{\num{0.79}}/\num{0.56}  \\ 
			Ours (Robust ICP-\emph{l}) & \num{0.36} & \num{57}/\num{0.6}   & \num{0.20} & \num{0.88}/\num{0.64}   & \num{0.21} & \num{0.92}/\num{0.93}   & \num{0.18} & \num{0.82}/\num{0.98}   & \num{0.17} & \num{0.79}/\num{0.56}  \\ 
			GMM & \num{4.37} & \num{130}/\num{120}   & \num{3.52} & \num{29}/\num{30}   & \num{5.66} & \num{63}/\num{49}   & \num{4.02} & \num{19}/\num{12}   & \num{5.73} & \num{110}/\num{120}  \\ 
			CPD & \num{2053.87} & \num{59}/\num{75}   & \num{1115.53} & \num{17}/\num{9.9}   & \num{2646.56} & \num{18}/\num{13}   & \num{2056.87} & \num{13}/\num{8.8}   & \num{1729.61} & \num{34}/\num{32}  \\ 
			Teaser++ & \num{3.07} & \num{180}/\num{180}   & \num{2.80} & \num{38}/\num{20}   & \num{2.83} & \num{230}/\num{90}   & \num{2.75} & \num{230}/\num{320}   & \num{2.99} & \num{250}/\num{260}  \\ 
			DCP & \textbf{\num{0.02}} & \num{220}/\num{210}   & \textbf{\num{0.02}} & \num{320}/\num{360}   & \textbf{\num{0.02}} & \num{300}/\num{270}   & \textbf{\num{0.02}} & \num{330}/\num{350}   & \textbf{\num{0.02}} & \num{220}/\num{220}  \\ 
			DGR & \num{1.10} & \num{26}/\num{28}   & \num{1.12} & \num{6.5}/\num{5.7}   & \num{1.29} & \num{9.8}/\num{9.8}   & \num{0.94} & \num{6.5}/\num{5.2}   & \num{1.25} & \num{11}/\num{8.9} \\\Xhline{1pt}
		\end{tabular}
	}
\end{table*}

We further test the methods on 25 pairs of partially overlapping point clouds constructed from five models in~\citemain{ZhouPK16}, with five pairs for each model. 
Fig.~\ref{fig:FGR_pc} compares the results on some problem instances, showing their computational time and RMSE, and using color-coding to visualize the deviation from the ground-truth alignment.
Tab.~\ref{Tab:FRG_Time_MSE} shows the average computational time and average/median RMSE on each model for each method. 
Overall, our robust methods and sparse ICP lead to more accurate results. Our methods achieve best average/median RMSE measures in more instances,
while being significantly faster than sparse ICP.

\begin{figure}[!t]
	\centering
		\includegraphics[width=\columnwidth]{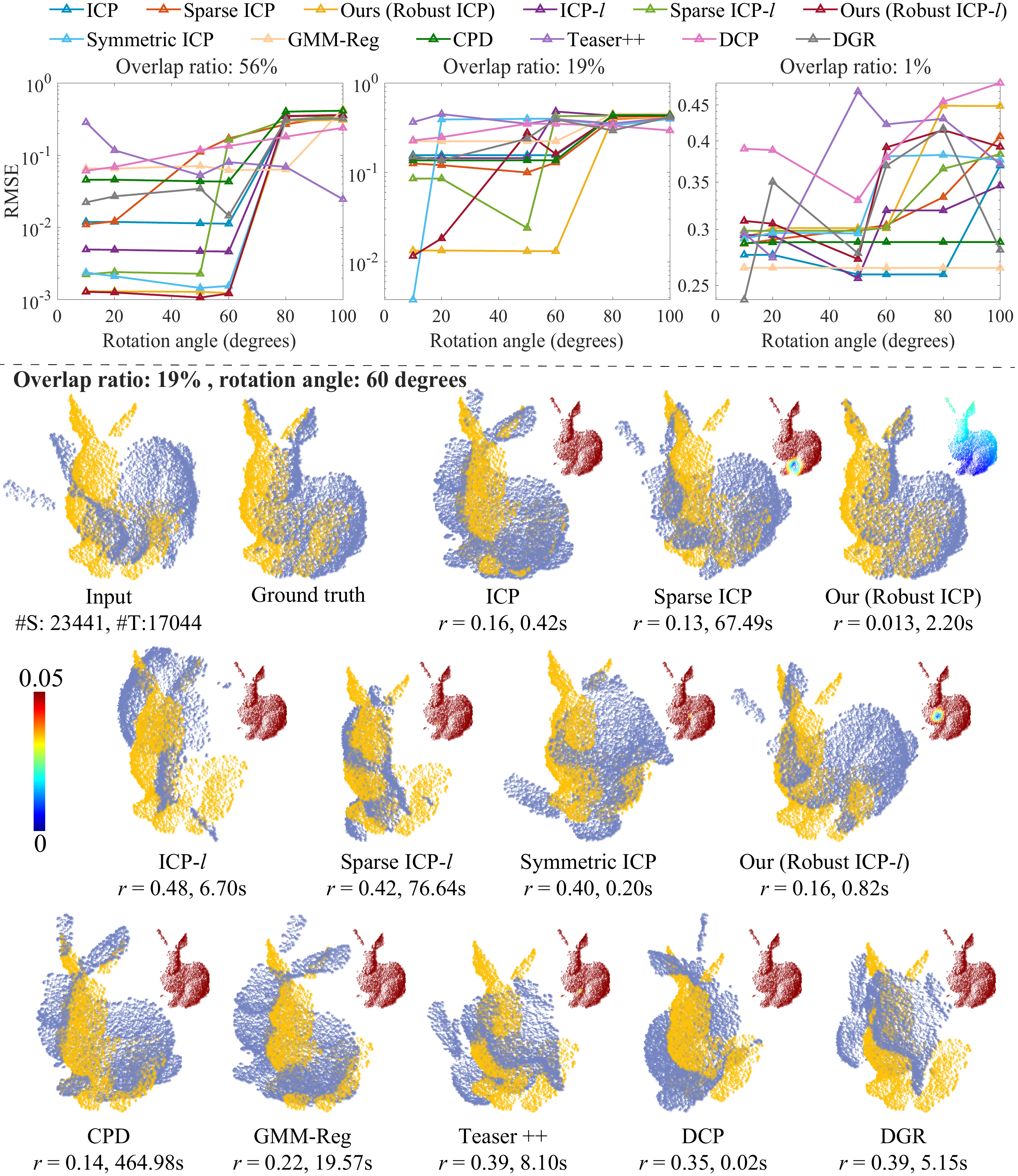}
		\caption{Registration results on simulated Kinect point clouds from the Stanfard bunny model, with different overlap ratios, and different amounts of rotation between the initial alignment to the ground truth.}	
	\label{fig:diff_overlap}
\end{figure}

In Fig.~\ref{fig:diff_overlap}, we evaluate how partial overlaps and initialization affect the registration accuracy of different methods. For the Stanford bunny model, we use the method from~\citemain{bohg2014robot} to simulate four point clouds captured using Kinect from different locations on the same horizontal plane as the model. We take one of the point clouds as the source, and each of the remaining three as the target for registration. 
For each pair of point clouds, we first place them according to their ground-truth alignment and perform PCA on the points, then rotate the target point cloud around the PCA axis with the smallest variance by an angle $\beta$ as initial alignment. As $\beta$ increases, the initialization deviates more from the ground-truth alignment. We test the methods with $\beta = 10^\circ$, $20^\circ$, $50^\circ$, $60^\circ$, $80^\circ$ and $100^\circ$, respectively. Fig.~\ref{fig:diff_overlap} plots the resulting RMSE values on each pair of point clouds with different values of $\beta$, together with the overlapping ratio with respect to the source point cloud. For all methods, the registration accuracy deteriorates as the overlap ratio decreases and the rotation angle increases. For an overlap ratio of $59\%$, our robust methods and symmetric ICP can achieve small RMSE values at the scale of $1 \times 10^{-3}$ with a rotation angle up to $60\%$. For an overlap ratio of $19\%$, our robust point-to-point ICP can still achieve an RMSE at the scale of $1 \times 10^{-2}$ with a rotation angle up to $60\%$, while other methods perform notably worse. With $1\%$ overlap, all methods result in large RMSE values regardless of the rotation angle.

\begin{table*}[!t]
	\caption{Average computational time (in seconds) and average/median RMSE ($\times 10^{-2}$) using different registration methods for eight sequences from the RGB-D SLAM dataset~\protect\citemain{SturmEEBC12}, with  best performance numbers highlighted in bold fonts.}
	\label{Tab:RGBD_Time_MSE}
	\setlength{\tabcolsep}{0.9pt}
	\centering
	{
		\fontsize{8}{9.6}\selectfont
		\begin{tabular}{ | c |  c  c | c  c | c  c | c  c  |  c c | c c | c c | c c |}
			\Xhline{1pt}
			\multicolumn{1}{ | c |}{\multirow{2}{*}{Dataset}}
			&\multicolumn{2}{ c |}{xyz}	
			&\multicolumn{2}{ c | }{360}
			&\multicolumn{2}{ c | }{teddy}
			&\multicolumn{2}{ c |}{desk}	
			&\multicolumn{2}{ c | }{plant} &\multicolumn{2}{ c |}{dishes}	
			&\multicolumn{2}{ c |}{coke}
			&\multicolumn{2}{ c | }{flowerbouquet}\\\cline{2-17}
			\multicolumn{1}{ | c |}{}
			&Time  	&RMSE	
			&Time 	 &RMSE	
			&Time  	&RMSE	
			&Time  	&RMSE
			&Time 	 &RMSE	
			&Time  	&RMSE	
			&Time  	&RMSE
			&Time 	 &RMSE \\\hline
			ICP & \num{0.23} & \num{2.1}/\num{0.89}   & \num{0.76} & \num{5.1}/\num{4}   & \num{0.68} & \num{2.1}/\num{1.4}   & \num{0.26} & \num{2.3}/\num{1.2}   & \num{0.51} & \num{1.6}/\num{1.1}   & \num{0.80} & \num{3.7}/\num{2.8}   & \num{0.93} & \num{3.1}/\num{2.5}   & \num{0.91} & \num{2.7}/\num{2.2}  \\ 
			AA-ICP & \num{0.16} & \num{2.1}/\num{0.9}   & \num{0.48} & \num{5.1}/\num{4}   & \num{0.38} & \num{2.1}/\num{1.4}   & \num{0.17} & \num{2.3}/\num{1.2}   & \num{0.32} & \num{1.6}/\num{1.1}   & \num{0.43} & \num{3.7}/\num{2.8}   & \num{0.51} & \num{3.1}/\num{2.5}   & \num{0.59} & \num{2.7}/\num{2.1}  \\ 
			Ours (Fast ICP) & \num{0.14} & \num{2.1}/\num{0.87}   & \num{0.36} & \num{5.1}/\num{4}   & \num{0.35} & \num{2.1}/\num{1.4}   & \num{0.15} & \num{2.3}/\num{1.2}   & \num{0.26} & \num{1.6}/\num{1.1}   & \num{0.37} & \num{3.7}/\num{2.8}   & \num{0.43} & \num{3.1}/\num{2.5}   & \num{0.43} & \num{2.7}/\num{2.2}  \\ 
			Sparse ICP & \num{11.2} & \num{1.6}/\num{0.86}   & \num{36.4} & \num{4.8}/\num{3.7}   & \num{51.7} & \num{1.8}/\num{1.1}   & \num{26.7} & \num{1.8}/\num{1.1}   & \num{71.9} & \num{0.88}/\num{0.67}   & \num{57.3} & \num{3.9}/\num{3.6}   & \num{77.1} & \num{3.3}/\num{3}   & \num{66.8} & \num{3.2}/\num{3}  \\ 
			Ours (Robust ICP) & \num{0.60} & \textbf{\num{0.5}}/\num{0.43}   & \num{2.86} & \num{2.2}/\num{0.75}   & \num{2.69} & \textbf{\num{1}}/\textbf{\num{0.76}}   & \num{0.93} & \num{1.2}/\num{0.77}   & \num{2.25} & \textbf{\num{0.65}}/\textbf{\num{0.56}}   & \num{2.36} & \num{3.2}/\num{2.6}   & \num{3.73} & \num{2.4}/\num{2.1}   & \num{3.52} & \num{2.4}/\num{1.8}  \\ 
			ICP-\emph{l} & \num{0.76} & \num{2.6}/\num{0.63}   & \num{1.93} & \num{4.3}/\num{2.2}   & \num{1.31} & \num{1.9}/\num{1.2}   & \num{0.66} & \num{14}/\num{0.9}   & \num{1.13} & \num{1.3}/\num{0.89}   & \num{1.28} & \num{11}/\num{2.7}   & \num{0.95} & \num{3.1}/\num{2.6}   & \num{1.00} & \num{2.7}/\num{2.1}  \\ 
			Sparse ICP-\emph{l} & \num{31.0} & \num{0.85}/\textbf{\num{0.43}}   & \num{64.0} & \num{2.6}/\num{0.86}   & \num{86.3} & \num{1.3}/\num{0.79}   & \num{33.1} & \textbf{\num{1.2}}/\textbf{\num{0.63}}   & \num{62.4} & \num{0.83}/\num{0.6}   & \num{65.6} & \num{790}/\num{3.6}   & \num{108} & \num{3.5}/\num{3}   & \num{92.2} & \num{3.4}/\num{3.3}  \\ 
			Symmetric ICP & \num{0.18} & \num{1.2}/\num{0.44}   & \num{0.36} & \num{1.8}/\textbf{\num{0.73}}   & \num{0.47} & \num{1.1}/\num{0.82}   & \num{0.19} & \num{1.7}/\num{0.7}   & \num{0.42} & \num{0.7}/\num{0.59}   & \num{0.39} & \num{3.8}/\num{3}   & \num{0.54} & \num{3.4}/\num{2.9}   & \num{0.53} & \num{3.3}/\num{3}  \\ 
			Ours (Robust ICP-\emph{l}) & \num{0.43} & \num{1.1}/\num{0.43}   & \num{1.25} & \num{2.6}/\num{0.84}   & \num{1.34} & \num{1.3}/\num{0.82}   & \num{0.56} & \num{1.5}/\num{0.64}   & \num{1.16} & \num{0.71}/\num{0.57}   & \num{1.11} & \num{3.8}/\num{3.4}   & \num{1.50} & \num{3.3}/\num{3}   & \num{1.46} & \num{3.2}/\num{3}  \\ 
			GMM-Reg & \num{1.88} & \num{4.4}/\num{3.7}   & \num{2.31} & \num{6.3}/\num{4.6}   & \num{1.63} & \num{3.2}/\num{2.6}   & \num{1.95} & \num{5.7}/\num{4.6}   & \num{1.72} & \num{2.7}/\num{2.1}   & \num{1.65} & \num{3.7}/\num{2.7}   & \num{1.73} & \num{3.1}/\num{2}   & \num{1.63} & \num{2.4}/\num{1.8}  \\ 
			CPD & \num{427} & \num{4.6}/\num{3.6}   & \num{423} & \num{5.3}/\num{3.3}   & \num{442} & \num{2.2}/\num{1.5}   & \num{348} & \num{2.1}/\num{1.8}   & \num{442} & \num{1.4}/\num{1.1}   & \num{434} & \num{3.4}/\num{2.7}   & \num{433} & \num{2.6}/\num{1.8}   & \num{430} & \num{2.3}/\num{1.8}  \\ 
			Teaser++ & \num{1.36} & \num{3.4}/\num{2}   & \num{1.05} & \num{20}/\num{14}   & \num{0.77} & \num{11}/\num{3.2}   & \num{5.89} & \num{8}/\num{2.5}   & \num{0.83} & \num{6.1}/\num{2.4}   & \num{0.73} & \num{21}/\num{15}   & \num{0.58} & \num{23}/\num{21}   & \num{0.59} & \num{20}/\num{11}  \\ 
			DCP & \textbf{\num{0.02}} & \num{6.5}/\num{5.4}   & \textbf{\num{0.02}} & \num{10}/\num{9.9}   & \textbf{\num{0.02}} & \num{6.6}/\num{5.5}   & \textbf{\num{0.02}} & \num{7}/\num{6}   & \textbf{\num{0.02}} & \num{5.6}/\num{4.9}   & \textbf{\num{0.02}} & \num{7.4}/\num{6}   & \textbf{\num{0.02}} & \num{7.5}/\num{6}   & \textbf{\num{0.02}} & \num{6.3}/\num{5.7}  \\ 
			DGR & \num{0.68} & \num{0.6}/\num{0.53}   & \num{1.14} & \textbf{\num{1.4}}/\num{0.86}   & \num{1.19} & \num{1}/\num{0.84}   & \num{0.77} & \num{1.2}/\num{0.96}   & \num{1.15} & \num{0.71}/\num{0.65}   & \num{3.54} & \textbf{\num{2.6}}/\textbf{\num{1.7}}   & \num{4.69} & \textbf{\num{2.1}}/\textbf{\num{1.2}}   & \num{3.96} & \textbf{\num{1.9}}/\textbf{\num{1.1}}  \\ \Xhline{1pt}
		\end{tabular}
	}
\end{table*}

\begin{figure*}[!t]
	\centering
	\includegraphics[width=2.0\columnwidth]{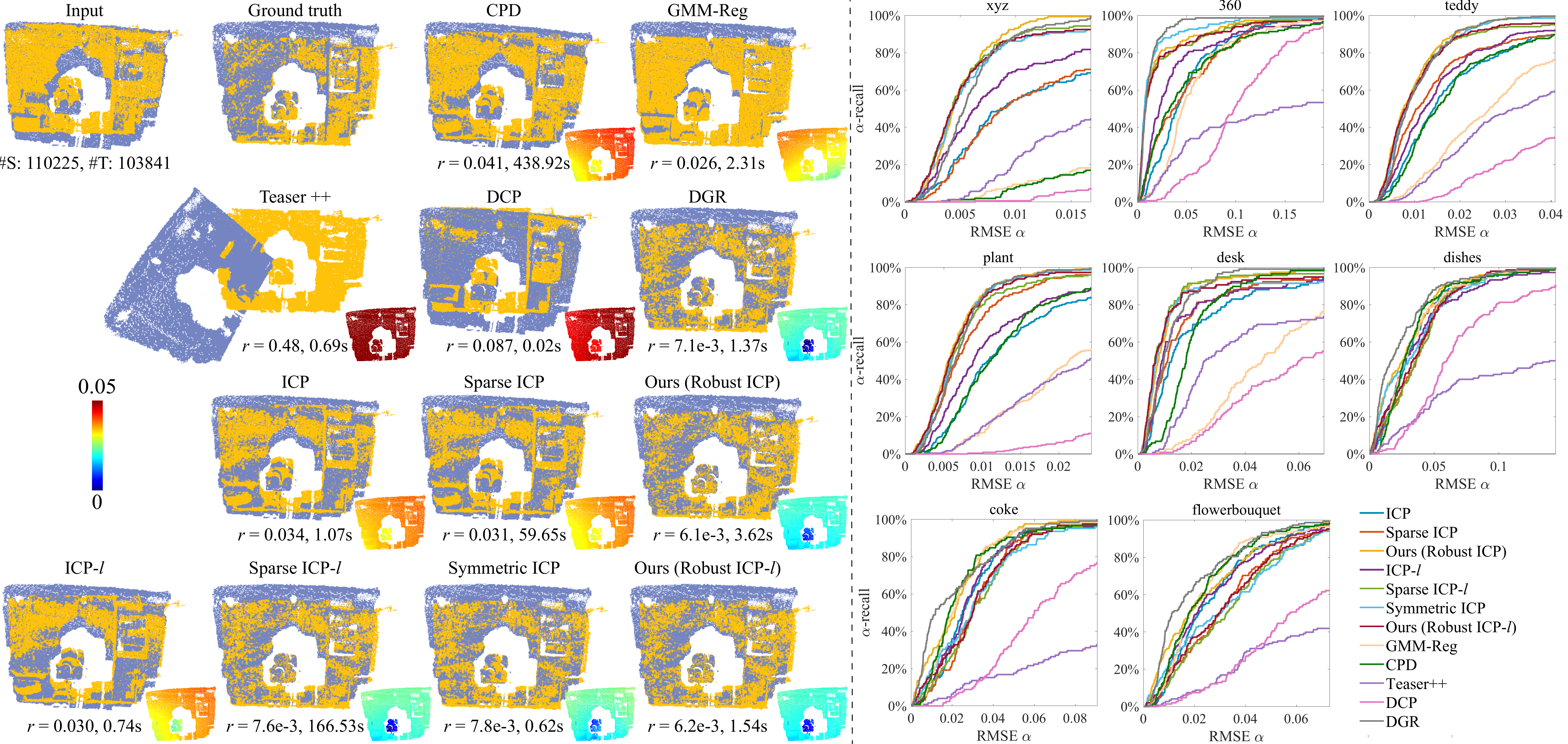}
	\caption{Examples of registration results using different methods on eight sequences from the RGB-D SLAM dataset~\protect\citemain{SturmEEBC12}, with color-coding to visualize the deviation from the ground-truth alignment. The plots on the right show the $\alpha$-recall rates of different methods for each point cloud sequence from the dataset.}
	\label{fig:RGBD}
\end{figure*}

\subsection{Real-World Data}
To evaluate their performance on real-world problems, we test the methods on the RGB-D SLAM dataset~\citemain{SturmEEBC12}, the ETH laser registration dataset~\citemain{Pomerleau2012}, and the 3DMatch dataset~\citemain{Zeng2017}. 
For the RGB-D SLAM dataset, We use eight  point cloud sequences captured with two cameras (``xyz'', ``360'',``teddy'', ``desk'' and ``plant'' for camera 1; `dishes'', ``coke'' and ``flowerbouquet'' for camera 2). For each sequence, we register pairs of point clouds that are a fixed number of frames apart (five frames for camera 1, and 20 frames for camera 2, taking into consideration the different velocities of the two cameras). As the two point clouds are already close to each other, we directly apply the registration methods without pre-alignment. 
For the ETH laser registration dataset, we test all of its eight point cloud sequences each containing between 31 and 45 point clouds, and we align each pair of adjacent point clouds from each sequence. 
For the 3DMatch dataset, we use the point cloud pairs in their geometric registration benchmark, and divide them into five categories according to the overlapping ratio with respect to the source point cloud: $[0\%, 20\%)$, $[20\%, 40\%)$, $[40\%, 60\%)$, $[60\%, 80\%)$, and $[80\%, 100\%]$. Within each category we sample 50 pairs to perform registration. 
All point clouds are pre-processed using a box grid filter to make the density more uniform, which is the same as the pre-processing operation in~\citemain{vongkulbhisal2018inverse}.

\begin{table*}[!t]
	\caption{Average computational time (in seconds) and average/median RMSE ($\times 10^{-3}$) for different registration methods on point cloud pairs in eight sequences from the ETH laser registration dataset~\protect\citemain{Pomerleau2012}. Best performance numbers are highlighted in bold fonts.}
	\label{Tab:ETH_Time_MSE}
	\setlength{\tabcolsep}{0.85pt}
	\centering
	{
		\fontsize{8}{9.6}\selectfont
		\begin{tabular}{ | c |  c  c | c  c | c  c | c  c  |  c c | c c | c c | c c |}
			\Xhline{1pt}
			\multicolumn{1}{ | c |}{\multirow{2}{*}{Method}}
			&\multicolumn{2}{ c |}{Apartment}	
			&\multicolumn{2}{ c | }{\tabincell{c}{ETH\\ Hauptgebaude}}
			&\multicolumn{2}{ c | }{Stairs}
			&\multicolumn{2}{ c |}{Mountains}	
			&\multicolumn{2}{ c |}{\tabincell{c}{Gazebo \\in summer}}	
			&\multicolumn{2}{ c |}{\tabincell{c}{Gazebo \\in winter}}	
			&\multicolumn{2}{ c | }{\tabincell{c}{Wood \\in summer}}
			&\multicolumn{2}{ c | }{\tabincell{c}{Wood \\in winter}}\\\cline{2-17}
			\multicolumn{1}{ | c |}{}
			&Time  	&RMSE	
			&Time 	 &RMSE	
			&Time  	&RMSE	
			&Time  	&RMSE
			&Time 	 &RMSE	
			&Time  	&RMSE	
			&Time  	&RMSE
			&Time 	 &RMSE \\\hline
			ICP & \num{0.06} & \num{36}/\num{12}   & \num{0.23} & \num{33}/\num{2.5}   & \num{0.15} & \num{16}/\num{2.4}   & \num{0.12} & \num{12}/\num{6.1}   & \num{0.14} & \num{15}/\num{7.5}   & \num{0.17} & \num{14}/\num{1.7}   & \num{0.22} & \num{9.7}/\num{1.4}   & \num{0.16} & \num{8.8}/\num{1.3}  \\ 
			AA-ICP & \num{0.08} & \num{45}/\num{12}   & \num{0.24} & \num{35}/\num{2.6}   & \num{0.14} & \num{16}/\num{2.4}   & \num{0.12} & \num{12}/\num{6.2}   & \num{0.15} & \num{15}/\num{7.5}   & \num{0.21} & \num{14}/\num{1.7}   & \num{0.18} & \num{9.7}/\num{1.4}   & \num{0.25} & \num{8.8}/\num{1.3}  \\ 
			Ours (Fast ICP) & \num{0.05} & \num{36}/\num{12}   & \num{0.17} & \num{33}/\num{2.6}   & \num{0.08} & \num{17}/\num{2.6}   & \num{0.08} & \num{12}/\num{6.2}   & \num{0.14} & \num{15}/\num{7.5}   & \num{0.14} & \num{14}/\num{1.7}   & \num{0.23} & \num{9.7}/\num{1.4}   & \num{0.15} & \num{8.8}/\num{1.3}  \\ 
			Sparse ICP & \num{1.23} & \num{13}/\num{2}   & \num{13.6} & \num{39}/\num{10}   & \num{2.33} & \num{6}/\num{1.2}   & \num{4.22} & \num{4.7}/\num{0.62}   & \num{7.18} & \num{11}/\num{0.75}   & \num{12.8} & \num{17}/\num{0.51}   & \num{13.6} & \num{11}/\num{0.44}   & \num{13.4} & \num{7.2}/\num{0.46}  \\ 
			Ours (Robust ICP) & \num{0.24} & \num{13}/\num{1.9}   & \num{0.83} & \num{15}/\num{0.55}   & \num{0.25} & \num{5.6}/\num{1.2}   & \num{0.22} & \num{3}/\num{0.6}   & \num{0.40} & \num{0.77}/\num{0.72}   & \num{0.63} & \num{10}/\num{0.35}   & \num{0.72} & \num{8.3}/\textbf{\num{0.42}}   & \num{0.48} & \textbf{\num{0.4}}/\textbf{\num{0.36}}  \\ 
			ICP-\emph{l} & \num{0.74} & \num{41}/\num{7.1}   & \num{0.10} & \num{33}/\num{1.4}   & \num{0.36} & \num{8.3}/\num{1.2}   & \num{0.08} & \num{6}/\num{3.1}   & \num{0.28} & \num{12}/\num{6.5}   & \num{0.53} & \num{18}/\num{1.5}   & \num{0.34} & \num{14}/\num{1.5}   & \num{0.64} & \num{9.1}/\num{1.5}  \\ 
			Sparse ICP-\emph{l} & \num{11.0} & \num{12}/\num{0.83}   & \num{14.1} & \num{35}/\textbf{\num{0.41}}   & \num{14.0} & \num{4.8}/\textbf{\num{0.27}}   & \num{13.0} & \num{1.7}/\textbf{\num{0.59}}   & \num{10.3} & \textbf{\num{0.63}}/\num{0.58}   & \num{11.6} & \num{12}/\textbf{\num{0.29}}   & \num{17.7} & \num{11}/\num{0.44}   & \num{18.5} & \num{7.2}/\num{0.47}  \\ 
			Symmetric ICP & \num{0.09} & \num{11}/\num{1.1}   & \num{0.22} & \num{33}/\num{0.44}   & \num{0.10} & \num{6}/\num{0.49}   & \num{0.10} & \textbf{\num{0.84}}/\num{0.82}   & \num{0.17} & \num{6.4}/\num{0.67}   & \num{0.19} & \num{10}/\num{0.39}   & \num{0.24} & \num{8.5}/\num{0.65}   & \num{0.23} & \num{7.6}/\num{0.73}  \\ 
			Ours (Robust ICP-\emph{l}) & \num{0.13} & \textbf{\num{11}}/\textbf{\num{0.79}}   & \num{0.35} & \num{35}/\num{0.43}   & \num{0.17} & \num{4.8}/\num{0.32}   & \num{0.15} & \num{4.3}/\num{0.63}   & \num{0.27} & \num{2.6}/\textbf{\num{0.56}}   & \num{0.34} & \num{18}/\num{0.37}   & \num{0.48} & \num{11}/\num{0.43}   & \num{0.48} & \num{7.2}/\num{0.38}  \\ 
			GMM-Reg & \num{1.95} & \num{28}/\num{15}   & \num{1.78} & \num{43}/\num{9.7}   & \num{2.27} & \num{20}/\num{8.6}   & \num{2.46} & \num{15}/\num{11}   & \num{1.45} & \num{12}/\num{10}   & \num{1.58} & \num{20}/\num{7.9}   & \num{1.19} & \num{22}/\num{11}   & \num{1.13} & \num{19}/\num{11}  \\ 
			CPD & \num{127} & \num{28}/\num{6.4}   & \num{465} & \num{43}/\num{12}   & \num{246} & \num{8.2}/\num{3.5}   & \num{150} & \num{9.6}/\num{5.7}   & \num{454} & \num{12}/\num{8.5}   & \num{453} & \num{20}/\num{8}   & \num{458} & \num{23}/\num{13}   & \num{456} & \num{19}/\num{13}  \\ 
			Teaser++ & \num{0.53} & \num{260}/\num{310}   & \num{0.46} & \num{190}/\num{180}   & \num{0.48} & \num{150}/\num{130}   & \num{0.45} & \num{190}/\num{190}   & \num{0.45} & \num{230}/\num{250}   & \num{0.45} & \num{140}/\num{160}   & \num{0.47} & \num{180}/\num{180}   & \num{0.45} & \num{200}/\num{190}  \\ 
			DCP & \textbf{\num{0.02}} & \num{100}/\num{100}   & \textbf{\num{0.02}} & \num{20}/\num{15}   & \textbf{\num{0.02}} & \num{44}/\num{39}   & \textbf{\num{0.02}} & \num{40}/\num{34}   & \textbf{\num{0.02}} & \num{58}/\num{50}   & \textbf{\num{0.02}} & \num{32}/\num{27}   & \textbf{\num{0.02}} & \num{28}/\num{19}   & \textbf{\num{0.02}} & \num{45}/\num{32}  \\ 
			DGR & \num{5.85} & \num{15}/\num{1.7}   & \num{27.28} & \textbf{\num{7.4}}/\num{8.6}   & \num{0.61} & \textbf{\num{3.5}}/\num{1.8}   & \num{0.78} & \num{15}/\num{6}   & \num{0.87} & \num{1.7}/\num{1.1}   & \num{1.05} & \textbf{\num{0.6}}/\num{0.55}   & \num{1.37} & \textbf{\num{1}}/\num{0.94}   & \num{1.30} & \num{2.5}/\num{0.81}  \\ 
			\Xhline{1pt}
		\end{tabular}
	}
\end{table*}

\begin{figure*}[!t]
	\centering
	\includegraphics[width=\textwidth]{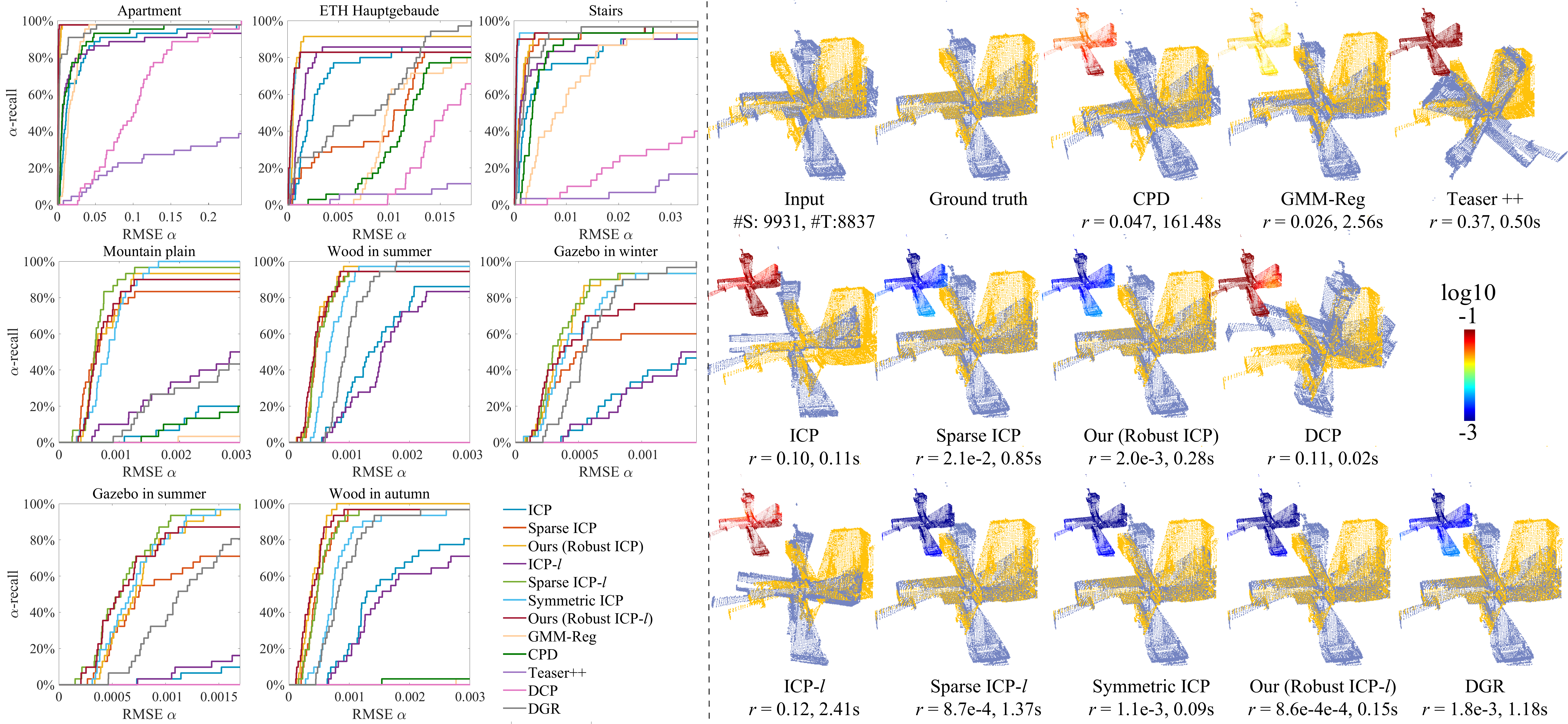}
	\caption{Examples of registration results with different methods on the ETH laser registration dataset~\protect\citemain{Pomerleau2012}, with color-coding for the deviation from the ground-truth alignment. The plots on the left show the $\alpha$-recall rates of each method for the point cloud sequences in the dataset.}
	\label{fig:ETH}
\end{figure*}

\begin{table*}[!ht]
	\caption{Average computational time (in seconds) and average/median RMSE ($\times 10^{-2}$) using different registration methods for the 3DMatch dataset~\protect\citemain{Zeng2017}, with  best performance numbers highlighted in bold fonts.}
	\label{Tab:3d_match}
	\setlength{\tabcolsep}{1.6pt}
	\centering
	\begin{footnotesize}
		\begin{tabular}{ | c |  c  c | c  c | c  c | c  c  |  c c | c c | }
			\Xhline{1pt}
			\multicolumn{1}{ | c |}{\multirow{2}{*}{Overlap}}
			&\multicolumn{2}{ c |}{0-20$\%$}	
			&\multicolumn{2}{ c | }{20$\%$-40$\%$}
			&\multicolumn{2}{ c | }{40$\%$-60$\%$}
			&\multicolumn{2}{ c |}{60$\%$-80$\%$}	
			&\multicolumn{2}{ c | }{80$\%$-100$\%$}\\\cline{2-11}
			\multicolumn{1}{ | c |}{}
			&Time  	&RMSE	
			&Time 	 &RMSE	
			&Time  	&RMSE	
			&Time  	&RMSE
			&Time 	 &RMSE \\\hline
			ICP & \num{0.30} & \num{43}/\num{42}   & \num{0.28} & \num{26}/\num{23}   & \num{0.21} & \num{16}/\num{11}   & \num{0.16} & \num{6.4}/\num{4}   & \num{0.06} & \num{1.8}/\num{1.1}  \\ 
			AA-ICP & \num{0.25} & \num{43}/\num{42}   & \num{0.27} & \num{26}/\num{23}   & \num{0.22} & \num{16}/\num{11}   & \num{0.13} & \num{6.4}/\num{4}   & \num{0.06} & \num{1.8}/\num{1.1}  \\ 
			Ours (Fast ICP) & \num{0.19} & \num{43}/\num{42}   & \num{0.16} & \num{26}/\num{23}   & \num{0.12} & \num{16}/\num{11}   & \num{0.09} & \num{6.4}/\num{4}   & \num{0.05} & \num{1.8}/\num{1.1}  \\ 
			Sparse ICP & \num{45.92} & \num{41}/\num{43}   & \num{42.54} & \num{22}/\num{18}   & \num{29.20} & \num{10}/\num{2.5}   & \num{20.38} & \num{3.1}/\num{0.74}   & \num{9.43} & \num{0.75}/\num{0.34}  \\ 
			Ours (Robust ICP) & \num{2.18} & \num{43}/\num{44}   & \num{1.27} & \num{22}/\num{19}   & \num{0.69} & \num{9.2}/\num{1}   & \num{0.38} & \num{2.5}/\textbf{\num{0.51}}   & \num{0.17} & \num{0.76}/\textbf{\num{0.33}}  \\ 
			ICP-\emph{l} & \num{3.91} & \num{44}/\num{44}   & \num{2.67} & \num{25}/\num{23}   & \num{1.53} & \num{14}/\num{7.9}   & \num{0.87} & \num{4.7}/\num{2}   & \num{0.13} & \num{1.4}/\num{0.69}  \\ 
			Sparse ICP-\emph{l} & \num{73.05} & \num{43}/\num{45}   & \num{67.49} & \num{21}/\num{16}   & \num{53.82} & \num{10}/\num{1.4}   & \num{35.63} & \num{2.6}/\num{0.54}   & \num{26.19} & \num{0.7}/\num{0.33}  \\ 
			Symmetric ICP & \num{0.17} & \num{40}/\num{45}   & \num{0.15} & \num{16}/\num{2.6}   & \num{0.12} & \num{9}/\textbf{\num{0.91}}   & \num{0.11} & \num{2.7}/\num{0.59}   & \num{0.09} & \textbf{\num{0.36}}/\num{0.34}  \\ 
			Ours (Robust ICP-\emph{l}) & \num{0.59} & \num{42}/\num{43}   & \num{0.40} & \num{20}/\num{15}   & \num{0.29} & \num{9.8}/\num{1.3}   & \num{0.22} & \num{2.5}/\num{0.56}   & \num{0.15} & \num{0.71}/\num{0.35}  \\ 
			GMM-Reg & \num{2.08} & \num{52}/\num{50}   & \num{2.33} & \num{33}/\num{32}   & \num{2.25} & \num{20}/\num{12}   & \num{2.01} & \num{11}/\num{7.2}   & \num{1.74} & \num{8.2}/\num{6.9}  \\ 
			CPD & \num{308.72} & \num{43}/\num{45}   & \num{390.49} & \num{30}/\num{27}   & \num{361.42} & \num{20}/\num{16}   & \num{291.68} & \num{9.4}/\num{6.2}   & \num{181.80} & \num{6.1}/\num{4.6}  \\ 
			Teaser++ & \textbf{\num{0.001}} & \num{50}/\num{52}   & \textbf{\num{0.001}} & \num{42}/\num{44}   & \textbf{\num{0.001}} & \num{34}/\num{38}   & \textbf{\num{0.001}} & \num{20}/\num{18}   & \textbf{\num{0.001}} & \num{18}/\num{14}  \\ 
			DCP & \num{0.02} & \num{37}/\num{38}   & \num{0.02} & \num{29}/\num{28}   & \num{0.02} & \num{22}/\num{20}   & \num{0.02} & \num{13}/\num{13}   & \num{0.02} & \num{8.6}/\num{8.3}  \\ 
			DGR & \num{2.94} & \textbf{\num{14}}/\textbf{\num{2.7}}   & \num{2.61} & \textbf{\num{1.5}}/\textbf{\num{1.3}}   & \num{1.33} & \textbf{\num{2.7}}/\num{0.96}   & \num{1.12} & \textbf{\num{0.75}}/\num{0.64}   & \num{0.86} & \num{0.51}/\num{0.43}  \\ 
			\Xhline{1pt}
		\end{tabular}
	\end{footnotesize}
\end{table*}

\begin{figure*}[!t]
	\centering
	\includegraphics[width=\textwidth]{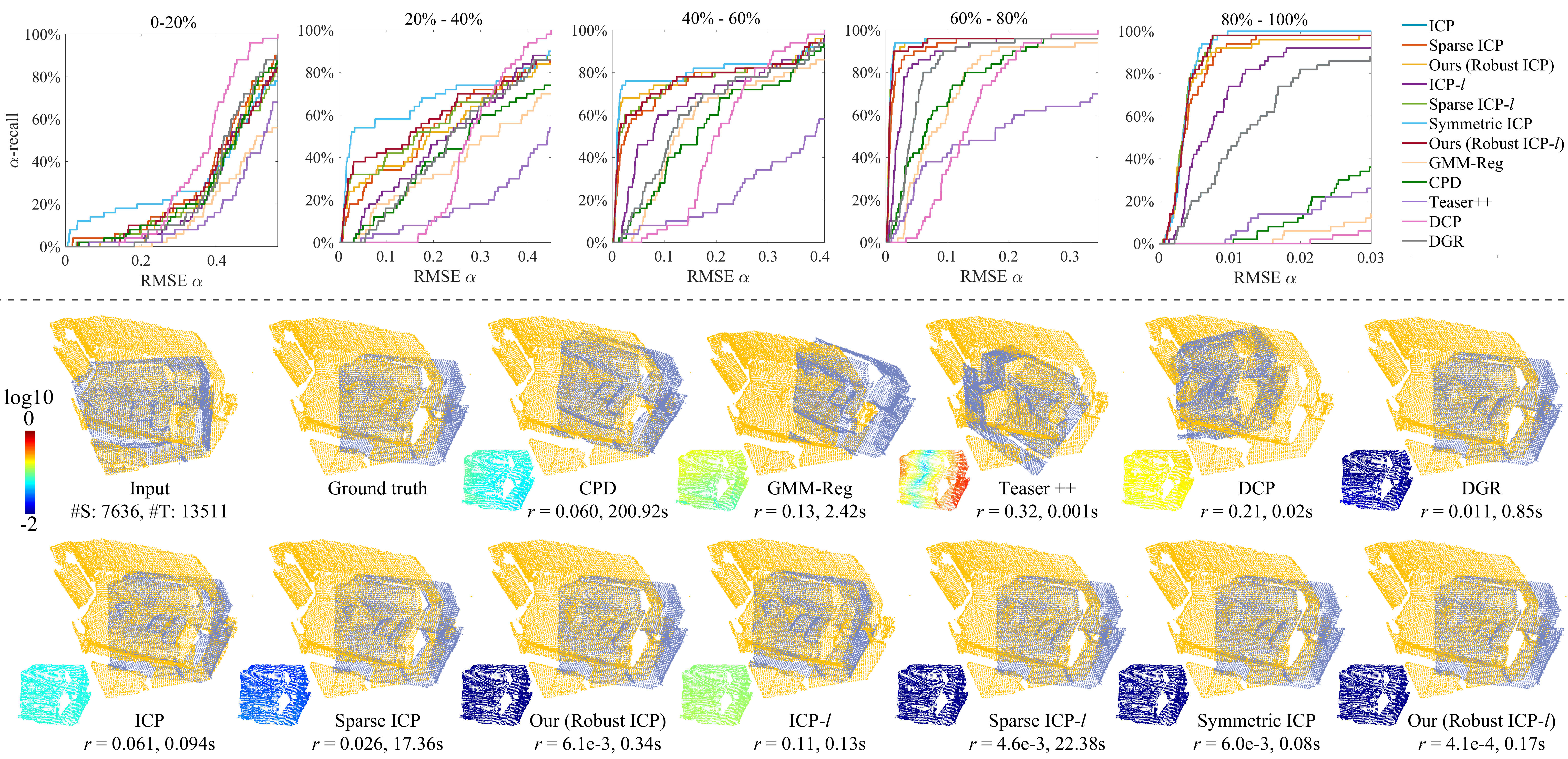}
	\caption{Examples of registration results for point clouds from the 3DMatch dataset~\protect\citemain{Zeng2017}, with color-coding to visualize the deviation from the ground-truth alignment. The plots on the top show the $\alpha$-recall rates of different methods on point cloud pairs in each range of overlapping ratio.}
	\label{fig:3dmatch}
\end{figure*}

\begin{figure*}[!t]
	\centering
	\includegraphics[width=\textwidth]{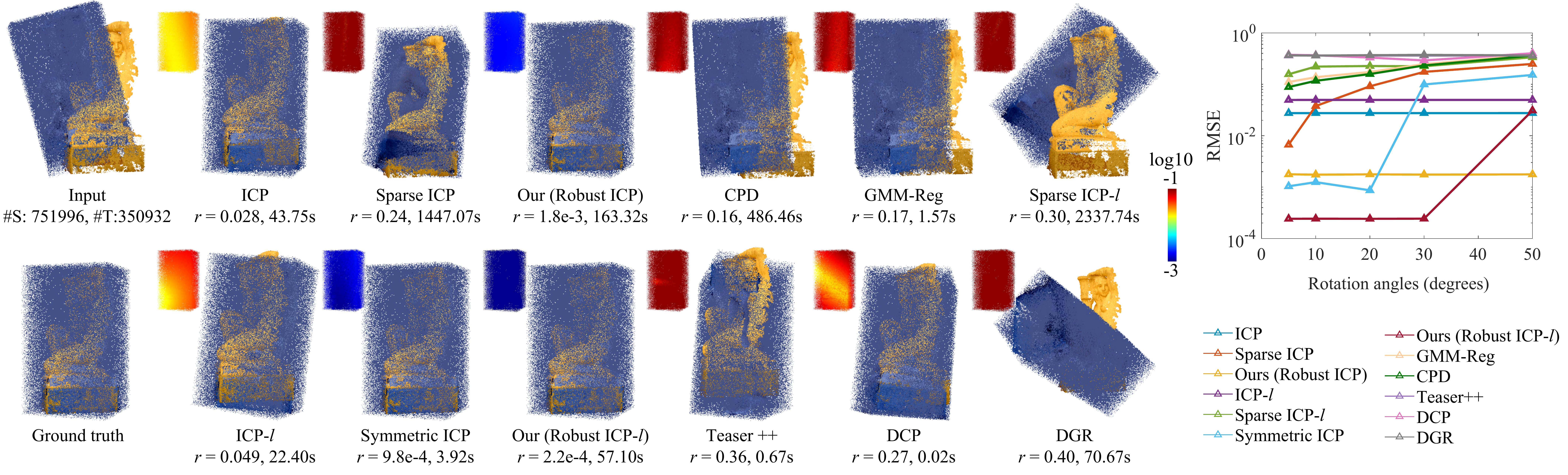}
	\caption{Registration results for the failure case in Fig.~\ref{fig:aquarius_noise} with 50\% added outliers, using random initial alignments that rotates the source point cloud from its ground-truth position by a fixed angle around a random axis. The plots on the right shows the average RMSE values for each rotation angle.}
	\label{fig:RandomInitialization}
\end{figure*}

Tables~\ref{Tab:RGBD_Time_MSE}, \ref{Tab:ETH_Time_MSE} and \ref{Tab:3d_match} show the average computational time and average/median RMSE for each method on the datasets, whereas Figures~\ref{fig:RGBD}, \ref{fig:ETH} and \ref{fig:3dmatch} show  examples of registration results together with color-coding of their deviation from the ground truth.  
To visualize the distribution of RMSE within each dataset, we also compute the \emph{$\alpha$-recall rate} $|S_{\alpha}| / |S|$ for each method, where $|S|$ is the total number of test cases, and $|S_{\alpha}|$ is the number of test cases where the RMSE is less than $\alpha$~\citemain{ZhouPK16}. 
Intuitively, for a given $\alpha$, a higher $\alpha$-recall rate indicates more test cases with RMSE values lower than $\alpha$. 
The plots of $\alpha$-recall rates for each method are included in Figures~\ref{fig:RGBD}, \ref{fig:ETH} and \ref{fig:3dmatch}.
For the RGB-D SLAM dataset and the ETH laser registration dataset, the majority of the lowest average/median RMSE values are achieved by our robust methods, sparse ICP and symmetric ICP. Like previous examples, our methods achieve similar or better accuracy than sparse ICP with much lower computational cost.
DGR has good performance on both datasets: on the RGB-D SLAM dataset it achieves the lowest average and median RMSE for all the camera-2 sequences, whereas on the ETH laser registration dataset it achieves the lowest average RMSE on many sequences. This is potentially due to  similar characteristics between its training data and the test cases.
For the 3DMatch dataset, ICP-based methods perform better on problems with overlap ratios higher than 40\%, with our robust point-to-point ICP and symmetric ICP attaining four out of the six lowest average/median RMSE values. For lower overlap ratios, DGR achieves the best accuracy because it is trained using the training set of the 3DMatch dataset and learns the characteristics of the test cases.

\subsection{Limitations}
\label{sec:limitations}
Like other ICP-based methods, our robust methods rely on good initial alignment. As shown in Fig.~\ref{fig:aquarius_noise} and Tab.~\ref{Tab:3d_match}, for some challenging problems, our methods may perform poorly because the initial alignment from Super4PCS deviates significantly from the ground truth.
In Fig.~\ref{fig:RandomInitialization}, we conduct another experiment for the failure case in Fig.~\ref{fig:aquarius_noise} with $50\%$ added outliers, using random initial alignments instead of Super4PCS. Specifically, we first rotate the source point cloud from its ground-truth position by a fixed angle $\beta$ around a random axis, and then perform registration. We test the methods with $\beta = 5^\circ$, $10^\circ$, $20^\circ$, $30^\circ$ and $50^\circ$, respectively. For each value of $\beta$, we conduct the experiment 10 times to construct 10 random initial alignments, and compute the average RMSE for each method. Fig.~\ref{fig:RandomInitialization} shows that our robust methods can still produce good results for such a challenging case if the initialization is not too far away from the ground truth. In particular, our robust point-to-plane ICP produces an average RMSE at the scale of $1 \times 10^{-4}$ with a rotation angle up to $30^\circ$, whereas our robust point-to-point ICP produces an average RMSE at the scale of $1 \times 10^{-3}$ with a rotation angle up to $50^\circ$. It verifies that the poor performance of our methods in Fig.~\ref{fig:aquarius_noise} is due to initialization.

For point clouds with a very small overlap, our methods may produce an incorrect result even with a good initial alignment (e.g., see Fig.~\ref{fig:diff_overlap}). This is partly due to our choice of the parameter $\nu$. Its initial value $\nu_{\max}$ is chosen based on the median initial alignment error, which is affected by the 50\% of source points that are closest to the target point cloud. If the proportion of source points in the overlapping region is significantly less than 50\%, then $\nu_{\max}$ may be much larger than the true initial distance. This may include too many source points into the initial iterations of the solver and lead it towards an incorrect result. 
 
\section{Conclusion and Future Work}
We proposed methods to improve the convergence speed and robustness of point-to-point and point-to-plane ICP methods. 
We first propose an Anderson-accelerated point-to-point ICP based on Lie algebra parameterization of rigid transformations, together with a stabilization strategy that ensures monotonic decrease of target energy. 
We also develop a robustified point-to-point ICP formulation based on the Welsch's function, and solve it using an Anderson-accelerated MM solver.
Finally, we extend the robust formulation and the accelerated numerical solver to point-to-plane ICP.
The resulting robust ICP schemes achieve similar or better accuracy than sparse ICP, while being significantly faster. The methods provide efficient and robust solutions to rigid registration problems where the data may be noisy, contain outliers, and overlap partially.

Our methods can be further improved in a few directions. 
First, to obtain good initial alignment for challenging cases, we can potentially adopt a machine learning-based method for determining a coarse alignment; this is similar to the practice in~\citemain{Wang2019} that uses ICP to refine a DCP alignment.
Second, we need a more sophisticated way to control the $\nu$ parameter and make our solver more robust on point clouds with a very small overlap; a data-driven approach could be a potential solution.
Finally, symmetric ICP shows promising performance in many of our comparisons; one interesting future work is to extend our approach to the symmetric ICP formulation, e.g. by replacing their $\ell_2$ target function with a robust error metric.

\section*{Acknowledgments}
This work was supported by National Natural Science Foundation of China (No. 61672481), and Youth Innovation Promotion Association CAS (No. 2018495). We thank Neil Gatenby for his help in proofreading.

\bibliographystyle{IEEEtran}
\bibliography{FastRobustRegistration}

\appendices

\section{Computing Matrix Logarithms}
\label{sec:MatrixLog}
To compute matrix logarithms, Existing numerical methods such as the \emph{inverse scaling and squaring method}~\citeappx{Higham2008} requires the matrix to have no negative eigenvalues, which may not hold for the transformation matrices considered in this paper. In the following, we derive a numerical method for computing logarithms of transformation matrices without such restrictions.

Given a transformation matrix
\[\mathbf{T}=\begin{bmatrix}
\mathbf{R} & \mathbf{t} \\
\mathbf{0} & 1
\end{bmatrix} \in \mathbb{R}^{(n+1) \times (n+1)} \]
where $\mathbf{R}$ is a rotation matrix, it can be shown that its real Schur decomposition has the following form~\citeappx{Gene1983,horn2012matrix}:
\begin{equation*}
\mathbf{T}=\mathbf{QUQ}^{T}
=\mathbf{Q}\left[
\begin{array}{ccccc}
{\mathbf{U}_{11}} & {\mathbf{U}_{12}} & {\mathbf{U}_{13}} & {\dots} & {\mathbf{U}_{1 m}} \\
{\mathbf{0}} & {\mathbf{U}_{22}} & {\mathbf{U}_{23}} & {\dots} & {\mathbf{U}_{2 m}} \\
{\vdots} & {\vdots} & {\vdots} & {\ddots} & {\vdots} \\
{\mathbf{0}} & {\mathbf{0}} & {\mathbf{0}} & {\dots} & {\mathbf{U}_{m m}}
\end{array}
\right] \mathbf{Q}^{T},
\end{equation*}
where $\mathbf{Q}\in\mathbb{R}^{(n+1)\times (n+1)}$ is an orthogonal matrix, and each diagonal block $\mathbf{U}_{ii}$ is either a 1-by-1 matrix or a 2-by-2 matrix. Due to the special form of $\mathbf{T}$, we can permute the rows and columns of $\mathbf{U}$ as well as the columns of $\mathbf{Q}$ to obtain the following decomposition:
\begin{equation}
\mathbf{T} = \mathbf{Q'} \mathbf{U'} \mathbf{Q'}^T
\label{eq:DiagonalSchur}
\end{equation}
where
\[\mathbf{Q'} =
\begin{bmatrix}
\mathbf{Q}_1 & 0 \\
0 & 1
\end{bmatrix},
\quad
\mathbf{U'} =
\begin{bmatrix}
\mathbf{D} & \mathbf{y} \\
\mathbf{0} & 1
\end{bmatrix},\]
and $\mathbf{D}$ is a block diagonal matrix~\citeappx{horn2012matrix}:
\begin{equation*}
\mathbf{D} =
\begin{bmatrix}
\mathbf{D}_1 &       &          \\
& \ddots &            \\
&      & \mathbf{D}_p       &   \\
&      & & \mathbf{I}_{n-2p}
\end{bmatrix}.
\end{equation*}
Here $\mathbf{D}_i$ is 2-by-2 rotation matrix and can be written as
\begin{equation}
\mathbf{D}_i =
\begin{bmatrix}
\cos{\theta_i} & -\sin{\theta_i} \\
\sin{\theta_i}  & \cos{\theta_i}
\end{bmatrix}
\label{eq:Di}
\end{equation}
with $\theta \in [0, \pi]$.
Using the decomposition~\eqref{eq:DiagonalSchur}, the logarithm of $\mathbf{T}$ can be computed as~\citeappx{Gene1983,horn2012matrix}:
\[
\log{(\mathbf{T})} = \log(\mathbf{Q'}\mathbf{U'}\mathbf{Q'}^T) = \mathbf{Q'}\log(\mathbf{U'})\mathbf{Q'}^T.
\]
To compute $\log(\mathbf{U'})$, we first note that
the rotation angle $\theta_i$ in~\eqref{eq:Di} can be determined from the entries of $\mathbf{D}_i$. We can then calculate the logarithm of $\mathbf{D}_i$ as
\begin{equation}
\label{eq:log_Di}
\mathbf{B}_i = \log(\mathbf{D}_i) =
\begin{bmatrix}
0 & -\theta_i \\
\theta_i  & 0
\end{bmatrix},
\end{equation}
Then we compute $\log(\mathbf{U'})$ according to~\citeappx{Gallier2002Computing} as
\begin{equation}
\label{eq:log_U}
\log{(\mathbf{U'})} = \log{
	\begin{bmatrix}
	\mathbf{D} & \mathbf{y}\\
	\mathbf{0} & 1
	\end{bmatrix}}
= \begin{bmatrix}
\mathbf{B} & \mathbf{Vy}\\
\mathbf{0} & 0
\end{bmatrix},
\end{equation}
where
\begin{equation}
\label{eq:log_D}
\mathbf{B} = \log{(\mathbf{D})} =
\begin{bmatrix}
\mathbf{B}_1 &       &          \\
& \ddots &            \\
&      & \mathbf{B}_p       &   \\
&      & & \mathbf{0}_{n-2p}
\end{bmatrix},
\end{equation}
and
\begin{equation}
\label{eq:calc_V}
\mathbf{V} = \mathbf{I}_n + \sum_{i=1}^p\left(-\frac{\theta_i}{2}\mathbf{B}_i + (1-\frac{\theta_i\sin{\theta_i}}{2(1-\cos{\theta_i})})\mathbf{B}_i^2\right).
\end{equation}
Algorithm~\ref{alg:Log_Trans} provides the psuedo-code for computing $\log(\mathbf{T})$.
\begin{algorithm}[h]
	\KwIn{Transform matrix $\mathbf{T} \in \seg{d}$.}
	\KwOut{$\hat{\mathbf{T}}=\log(\mathbf{T})$.}
	\BlankLine
	Compute the real Schur decomposition $(\mathbf{Q}, \mathbf{U})$ of $\mathbf{T}$\;
	Rearrange $\mathbf{Q}, \mathbf{U}$ to obtain matrices $\mathbf{Q'}, \mathbf{U'}$ in Eq.~\eqref{eq:DiagonalSchur}\;
	Calculate $\log(\mathbf{U'})$ according to Eqs.~\eqref{eq:log_Di}--\eqref{eq:calc_V}\;
	$\mathbf{\hat{T}} = \mathbf{Q'}\log(\mathbf{U'})\mathbf{Q'}^T$\;
	\caption{Logarithm for matrices in $\seg{d}$.}
	\label{alg:Log_Trans}
\end{algorithm}

\section{Surrogate Function for Eq.~\eqref{eq:WelschRegistration}}
\label{appx:SurrogateProof}
In this section, we show that the function in Eq.~\eqref{eq:TargetSurrogateFunc} is a surrogate function of the target function in Eq.~\eqref{eq:WelschRegistration} at the current transformation $\mathbf{R}^{(k)}, \mathbf{t}^{(k)}$.
To simplify notation, we denote
$D_i = D_i(\mathbf{R}, \mathbf{t})$ and  $D_i^{(k)} = D_i(\mathbf{R}^{(k)}, \mathbf{t}^{(k)})$.
By definition, since $\chi_{\nu}\bigl(D_i \mid D_i^{(k)} \bigl)$ is a surrogate function for $\psi_{\nu}(D_i)$ at $D_i^{(k)}$, we have
\begin{equation}
\begin{aligned}
\psi_{\nu} \bigl(D_i^{(k)}\bigl) &= \chi_{\nu}\bigl(D_i^{(k)} \mid D_i^{(k)} \bigl),\\
\psi_{\nu}(D_i) &\leq \chi_{\nu}\bigl(D_i \mid D_i^{(k)} \bigl),\quad \forall~D_i \neq D_i^{(k)}.
\end{aligned}
\label{eq:SurrogateChi}
\end{equation}
Note that the function $x_i(\mathbf{R}, \mathbf{t}) = \|\mathbf{R} \sourcept_i + \mathbf{t} - \paringpt_i^{(k)}\|$ satisfies
\begin{equation}
D_i \leq x_i(\mathbf{R}, \mathbf{t}), \quad  D_i^{(k)} = x_i(\mathbf{R}^{(k)}, \mathbf{t}^{(k)}).
\label{eq:XiBound}
\end{equation}
Moreover, $\chi_{\nu}(x \mid y)$ is a monotonically increasing function on $x \in [0, +\infty)$, which together with Eqs.~\eqref{eq:SurrogateChi} and \eqref{eq:XiBound} means that
\begin{equation*}
\begin{aligned}
\psi_{\nu} \bigl(D_i^{(k)}\bigl) &= \chi_{\nu}\bigl(D_i^{(k)} \mid D_i^{(k)} \bigl)= \chi_{\nu}\bigl(x_i(\mathbf{R}^{(k)}, \mathbf{t}^{(k)}) \mid D_i^{(k)} \bigl), \\
\psi_{\nu}(D_i) &\leq \chi_{\nu}\bigl(D_i \mid D_i^{(k)} \bigl) \leq \chi_{\nu}\bigl(x_i(\mathbf{R}, \mathbf{t}) \mid D_i^{(k)} \bigl).
\end{aligned}
\end{equation*}
It shows that $\chi_{\nu}\bigl(x_i(\mathbf{R}, \mathbf{t}) \mid D_i^{(k)} \bigl)$ is a surrogate function for $\psi_{\nu}(D_i)$ at $D_i^{(k)}$. Then substituting each term $\psi_{\nu}(D_i)$ in Eq.~\eqref{eq:WelschRegistration} with $\chi_{\nu}\bigl(x_i(\mathbf{R}, \mathbf{t}) \mid D_i^{(k)} \bigl)$, we can see that  Eq.~\eqref{eq:TargetSurrogateFunc} is a surrogate function at $(\mathbf{R}^{(k)}, \mathbf{t}^{(k)})$.

\section{Choices of $\nu_{\min}$}
\label{appx:SettingNu}
In this section, we explain the rationale for our choices of $\nu_{\min}$ for our robust ICP methods.

For the point-to-point method, our intention is to set $\nu_{\min}$ large enough such that point pairs with deviation due to difference in sampling locations will be included in the alignment step. For ease of discussion, we first assume that the target set has uniform sampling density, and the source set is sampled from a triangulated surface using the target set as vertices.  Then within the overlapping region the distance from a point in $\mathbf{p} \in \sourceset$ to the set $\targetset$ can be up to $E/{\sqrt{3}}$ where ${E}$ is the distance between neighboring points within $\targetset$ (e.g., when $\mathbf{p}$ lies at the center of an equilateral triangle $T$ with edge length ${E}$ from the triangulation of $\targetset$). In order to include $\mathbf{p}$ and its closest point into the alignment step, $\nu_{\min}$ should be no smaller than $1/3$ of the distance between them due to the three-sigma rule, i.e., $\nu_{\min} \geq E/{(3\sqrt{3})}$. In practice, the sampling density of $\targetset$ may not be uniform. Therefore, we compute the representative distance ${\overline{E}}_{\targetset}$ between neighboring points in $\targetset$ as explained in Section~\ref{sec:robust}, and set $\nu_{\min} = {\overline{E}}_{\targetset}/{(3\sqrt{3})}$.

For the point-to-plane method, we first use the same assumption as the point-to-point method: the target set has uniform sampling density, and the source set is sampled from a surface triangulated from the target set. For a source point $\sourcept \in \sourceset$, suppose its closest point in $\targetset$ is $\targetpt$. 

Such a point should be included into the alignment process. Recall that a point $\sourcept_i \in \sourceset$ will be effectively excluded from the  alignment step if the distance between its current transformed position and the point set $\targetset$ is larger than $3 \nu$. Therefore, to ensure the points in the overlapping region are included for alignment, $\nu_{\min}$ should be no smaller than $\frac{E}{3 \sqrt{3}}$. In reality, the sampling density of the point sets may not be uniform, thus we adapt the above heuristics as follows. We first compute the median distance from each point $\sourcept_i \in \sourceset$ to its six nearest neighbors in $\sourceset$, and take the median $\overline{E}_{\sourceset}$ of these median distance values across $\sourceset$. In the same way, we compute a value $\overline{E}_{\targetset}$ for the set $\targetset$. Let $\mathbf{s} \in \targetset$ be the neighbor point of $\targetpt$ that is the farthest away from the tangent plane at $\targetpt$. Denote by $H_{\targetpt}(\cdot)$ the distance to the tangent plane at $\targetpt$. Then we must have $H_{\targetpt}(\sourcept) \leq \frac{1}{2}H_{\targetpt}(\mathbf{s})$, otherwise $\targetpt$ would not be the closest point to $\sourcept$ in $\targetset$. Then in order to include the pair $(\sourcept, \targetpt)$ into the alignment step, we can set $\nu_{\min} \geq \frac{1}{6} H_{\targetpt}(\mathbf{s}) \geq \frac{1}{3}H_{\targetpt}(\sourcept)$. To handle non-uniform sampling of $\targetset$, we compute the representative distance $\overline{H}_{\targetset}$ to a neighboring point's tangent plane in $\targetset$, and set  $\nu_{\min} = \overline{H}_{\targetset} / 6$.

\section{Calculation of Gradient $\mathbf{J}^{(k)}$ in Eq.~\eqref{eq:Linearization}}
\label{appx:Gradient}
In this section, we show how to calculate the gradient $\mathbf{J}^{(k)}$ at $\mathbf{T}^{(k)}$ in Eq.~\eqref{eq:Linearization}.
We denotes the actual variables $\liealgparam{\mathbf{T}}=[\boldsymbol{\delta}^T, \mathbf{u}^T]^T$, where $\boldsymbol{\delta}=[{\delta}_1, \delta_2, \delta_3]$, then the gradient of ${B}_i^{(k)}$ can be represented as
\[
\mathbf{J}_i^{(k)}=\left[\frac{\partial B_i^{(k)}}{\partial \boldsymbol{\delta}^T},
\frac{\partial B_i^{(k)}}{\partial \mathbf{u}^T}\right]^T,
\]
and
\[
\left\{\begin{aligned}
\frac{\partial B_i^{(k)}}{\partial {\delta}_j}&=\left(\frac{\partial B_i^{(k)}}{\partial \mathbf{R}} \circ \frac{\partial \mathbf{R}}{\partial {\delta}_j}\right)_{\text{sum}}
+ \left(\frac{\partial B_i^{(k)}}{\partial \mathbf{t}} \circ \frac{\partial \mathbf{t}}{\partial {\delta}_j}\right)_{\text{sum}},\\
\frac{\partial B_i^{(k)}}{\partial \mathbf{u}_j}&=\left(\frac{\partial B_i^{(k)}}{\partial \mathbf{t}} \circ \frac{\partial \mathbf{t}}{\partial \mathbf{u}_j}\right)_{\text{sum}},
\end{aligned}\right.\]
where $1\leq j\leq 3$, $\circ$ is element-wise multiplication of two matrices, and $(\cdot)_{\text{sum}}$ is to add up each element of the matrix.
According to Eq.\eqref{eq:ProxyPointToPlaneAlignment}, we can calculate
\[
\frac{\partial B_i^{(k)}}{\partial \mathbf{R}}=\paringnormal_i^{(k)}\sourcept_i^T
,\quad \frac{\partial B_i^{(k)}}{\partial \mathbf{t}}=\paringnormal_i^{(k)}.\]
Then we compute the derivative of $(\mathbf{R}, \mathbf{t})$ about $\boldsymbol{\delta}$ and $\mathbf{u}$. According to~\citeappx{RichardZS94}, defining
\[
\mathbf{a}^{\wedge}=\left[\begin{array}{ccc}
0 & -a_{3} & a_{2} \\
a_{3} & 0 & -a_{1} \\
-a_{2} & a_{1} & 0
\end{array}\right],
\]
where $\mathbf{a}=(a_1,a_2,a_3)^T$.
Let $\|\boldsymbol{\delta}\|$ denates the $\ell_2$-norm of $\boldsymbol{\delta}$. When $\|\boldsymbol{\delta}\|\neq 0$, according to~\citeappx{Gallego15}, the derivatives of rotation matrix $\mathbf{R}$ about $\boldsymbol{\delta}$ is
\[
\frac{\partial \mathbf{R}}{\partial \delta_{j}}=\frac{\delta_{j}\boldsymbol{\delta}^{\wedge}+\left(\boldsymbol{\delta} \times(\mathbf{I} -\mathbf{R}) \mathbf{e}_{j}\right)^{\wedge}}{\|\boldsymbol{\delta}\|^{2}} \mathbf{R},
\]
where $\mathbf{e}_j$ is the $j$-th vector of the standard basis in $\mathbb{R}^3$ and $\mathbf{I}$ is the identity matrix in $\mathbb{R}^{3\times3}$.
The translation in $se(3)$ can be represented as~\citeappx{RichardZS94}
\[\mathbf{t}=\left((\mathbf{R}-\mathbf{I})\mathbf{u}^{\wedge}\boldsymbol{\delta}+\boldsymbol{\delta}\boldsymbol{\delta}^T\mathbf{u}\right) /\|\boldsymbol{\delta}\|^2.\]
We can compute the derivatives of $\mathbf{t}$ about $\boldsymbol{\delta}$ is
\[
\frac{\partial\mathbf{t}}{\partial\boldsymbol{\delta}}=\frac{1}{\|\boldsymbol{\delta}\|^2}\left( \mathbf{M}+\boldsymbol{\delta}\mathbf{u}^T+\mathbf{D}\right)
-\frac{2}{\|\boldsymbol{\delta}\|^4}\mathbf{M}\boldsymbol{\delta}^T\boldsymbol{\delta}
\]
where
\[
\mathbf{M}=(\mathbf{R}-\mathbf{I})\mathbf{u}^{\wedge}+\boldsymbol{\delta}^T\mathbf{uI}
\]
and the $j$-th column of $\mathbf{D}$ is
\[
\mathbf{D}_j=\frac{\partial \mathbf{R}}{\partial \delta_{j}}\mathbf{u}^{\wedge}\boldsymbol{\delta}
\]
And the derivatives of $\mathbf{t}$ about $\mathbf{u}$ is
\[
\frac{\partial\mathbf{t}}{\partial\mathbf{u}}=\frac{1}{\|\boldsymbol{\delta}\|^2}\left((\mathbf{I}-\mathbf{R}) \boldsymbol{\delta}^{\wedge}+\boldsymbol{\delta}\boldsymbol{\delta}^T\right).
\]
When $\|\boldsymbol{\delta}\|=0$,
\[
\frac{\partial \mathbf{R}}{\partial \delta_{j}}=\mathbf{e}_j^{\wedge}, \quad
\mathbf{t}=\mathbf{u}, \quad
\frac{\partial \mathbf{t}}{\partial \delta_{j}}=\mathbf{0}, \quad
\frac{\partial\mathbf{t}}{\partial\mathbf{u}}=\mathbf{I}.
\]

\section{Settings of Experiments}
\label{appx:Settings}
We follow the default settings of the open-source implementations for each method, except for the following changes:
\begin{itemize}
  \item ICP, ICP-$l$ and AA-ICP: For a fair comparison, we use the same termination criteria as in Algorithm~\ref{alg:AA-Robust-ICP} for a fixed parameter $\nu$: we terminate the solver if it reaches the maximum number of iterations (1000), or $\|\Delta \mathbf{T}\|_F^2<10^{-5}$, where $\Delta \mathbf{T}$ denotes the difference between the transformation from two consecutive iterations.
  \item Sparse ICP and Sparse ICP-$l$: For Sparse ICP, we choose $p=0.8$ for the RGB-D SLAM dataset, and $p=0.4$ for other experiments. For Sparse ICP-$l$, we choose $p=0.4$ for all experiments.
  \item CPD: Due to the high computational cost of CPD, for any point cloud with more than 15000 points, we downsample it to 15000 points using farthest point sampling.
  \item GMM-Reg: The documentation of the implementation recommends downsampling a point cloud to 5000 points for better performance. Therefore, for any point cloud with more than 5000 points, we downsample it to 5000 points using farthest point sampling.
  \item Teaser++: The implementation incurs high memory consumption, and requires the source and target point clouds to have the same number of points. Therefore, we first use farthest point sampling to downsample 5000 points on any point cloud containing more than 5000 points. Afterwards, if the source and target point clouds contain different numbers of points, we downsample the point cloud with more points to the same number of points as the other.
  For synthesized data with a known noise level, we set the noise bound parameter according to the noise level.
  \item DCP: We train the model using the 10000 pairs of point clouds from the training set of the 3DMatch dataset. For both the training and test data, we downsample the point clouds to 1024 points using farthest point sampling.
  \item DGR: We use the two pre-tained models (trained with 3DMatch and KITTI, respectively) released by the authors to test our examples.
  For use the KITTI-based model on five outdoor datasets (Mountains, Gazebo in summer, Gazebo in winter, Wood in summer, Wood in winter) and a mixed dataset (Stairs) in the ETH dataset. For all other problems, we use the model trained with 3DMatch.
\end{itemize}
In Fig.~\ref{fig:teaser-monkeys}, we count the number of iterations for each method as follows. For CPD, we count the iterations of the EM algorithm. For GMM-Reg method, we count the number of times for constructing the objective function. For Teaser++, we count the iterations for calculating the rotation matrix. For others, we count the number of times for the updating the corresponding points.

\begin{IEEEbiography}[{\includegraphics[width=1in]{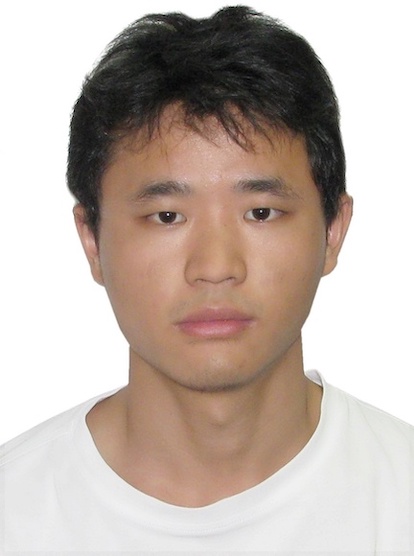}}]{Juyong Zhang}
is an associate professor in the School of Mathematical Sciences at University of Science and Technology of China. He received the BS degree from the University of Science and Technology of China in 2006, and the PhD degree from Nanyang Technological University, Singapore. His research interests include computer graphics, computer vision, and numerical optimization. He is an associate editor of The Visual Computer.
\end{IEEEbiography}

\begin{IEEEbiography}[{\includegraphics[width=1in]{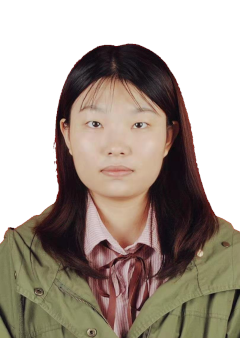}}]{Yuxin Yao} is currently working toward the master's degree in the School of Mathematical Sciences, University of Science and Technology of China. Her research interests include computer graphics and 3D registration.
\end{IEEEbiography}

\begin{IEEEbiography}[{\includegraphics[width=1in]{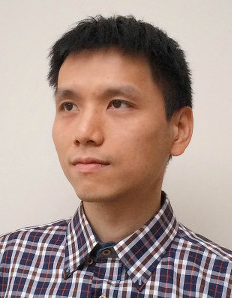}}]{Bailin Deng}
is a lecturer in the School of Computer Science and Informatics at Cardiff University. He received the BEng degree in computer software (2005) and the MSc degree in computer science (2008) from Tsinghua University (China), and the PhD degree in technical mathematics (2011) from Vienna University of Technology (Austria). His research interests include geometry processing, numerical optimization, computational design, and digital fabrication. He is a member of the IEEE.
\end{IEEEbiography}
\vfill

\end{document}